%% file: main.tex
\documentclass{article}

\usepackage{dilab_arxiv}

\usepackage{mathrsfs}
\usepackage{algorithm}
\usepackage{algorithmic}
\usepackage{listings}
\title{MLREF: Efficient Module Reuse for Reward Design in Reinforcement Learning via Large Language Models}
\runningtitle{MLREF: Efficient Module Reuse for Reward Design}
\date{arXiv preprint, \today}

\paperlogo{\includegraphics[height=1.5cm]{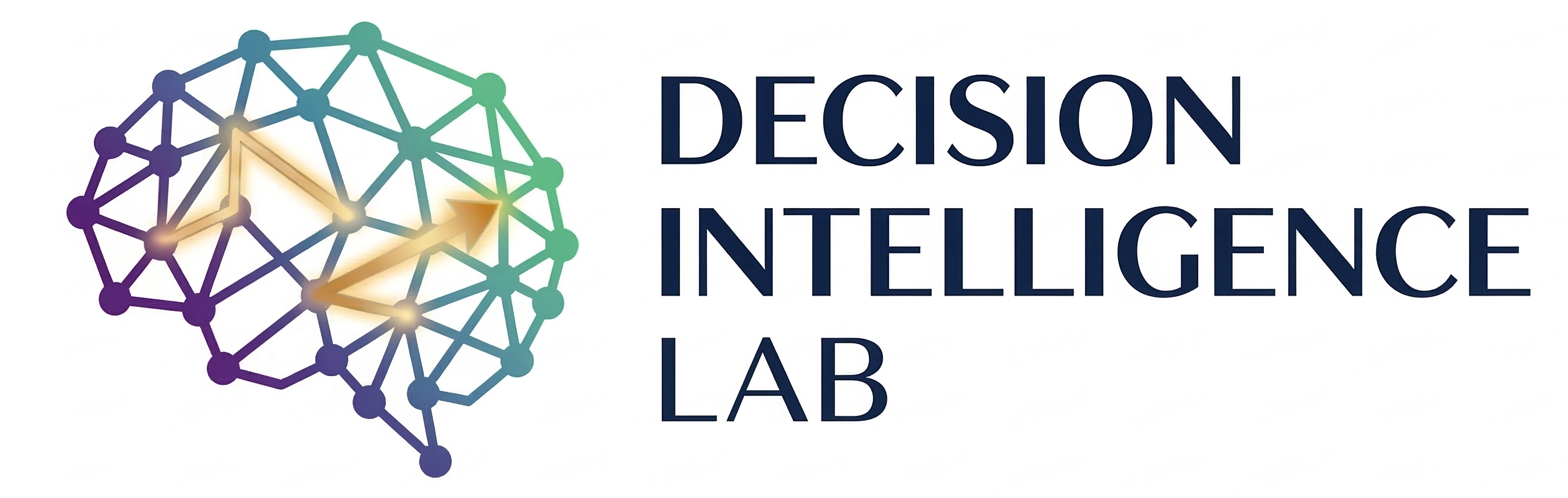}}

\author{
  Chenglin Liu$^{1}$, Xun Wang$^{1}$, Ruishuo Chen$^{1}$, Zhuoran Li$^{1}$, and Longbo Huang$^{1\,\text{\faEnvelope}}$
  \\[0.3em]\normalfont
  $^1$Institute for Interdisciplinary Information Sciences, Tsinghua University
  \\
  \text{\faEnvelope}\ Correspondence: longbohuang@tsinghua.edu.cn
}

\begin{document}

\maketitle
\thispagestyle{fancy}

\begin{abstract}
Reward function design remains a bottleneck in reinforcement learning. While large language models (LLMs) have enabled automated reward generation, existing methods generate and revise reward functions as monolithic programs, making it difficult to reliably preserve and reuse effective components discovered in earlier iterations, leading to unstable performance across iterations. To address this, we propose \textbf{M}odule \textbf{L}evel \textbf{R}eward \textbf{E}volution \textbf{F}ramework (MLREF). At the core of MLREF is a \emph{module pool}, a persistent repository of reusable reward components. MLREF treats the module pool as the primary optimization object: the pool evolves across iterations by accumulating successful modules, refining underperforming ones, and reusing proven components; while reward functions are constructed as linear combinations of modules drawn from this pool. To drive this evolution, MLREF integrates three mechanisms: reflection-based refinement, hybrid credit assignment, and a merge strategy with rollback, which together improve the effectiveness and robustness of reward optimization. Experiments on 17 tasks show that MLREF outperforms strong baselines by 25.2\% in locomotion and 6.6\% in manipulation, with more stable optimization dynamics.
\end{abstract}

\input{sections/01-introduction}
\input{sections/02-related_work}
\input{sections/03-preliminaries}
\input{sections/04-method}
\input{sections/05-experiments}
\input{sections/06-conclusion}

\bibliography{references}
\bibliographystyle{dilab_ref}

\makeappendixtoc
\appendix
\input{sections/appendix}

\end{document}

%% file: sections/01-introduction.tex
\section{Introduction}
\label{sec:intro}

Reinforcement learning (RL) has achieved remarkable success in various domains \cite{goldwaser2020deep, zhu2021deep, elguea2023review, radosavovic2024real}. 
However, its performance critically depends on the reward function, and designing an effective one remains a fundamental challenge. Sparse rewards are easy to define but often provide insufficient learning signals for optimization \cite{stanton2018deep, hare2019dealingsparserewardsreinforcement}, whereas dense rewards often require substantial domain expertise and manual engineering efforts \cite{Sutton+Barto:1998, Eschmann2021}.

{Recent advances in large language models (LLMs) have provided new opportunities for automating reward design, leveraging their strong capabilities in reasoning \cite{wei2022chain, zhang2024chain}, instruction following \cite{ouyang2022training}, and code generation \cite{chen2021evaluating}. However, directly generated rewards often suffer from hallucination, syntactic errors, and semantic misalignment with task objectives, limiting their reliability in practical RL scenarios \cite{ma2024eureka, sun2025large}. 
These challenges motivate the development of iterative refinement frameworks, where LLMs progressively improve reward functions based on feedback from RL training.}

To improve the quality of LLM-generated rewards, numerous LLM-based harnesses and workflows have been developed to iteratively refine reward functions through interaction with RL training feedback. However, existing approaches predominantly optimize rewards at the function level by generating or modifying monolithic reward programs, e.g., EUREKA \cite{ma2024eureka}, $R^{*}$ \cite{li2025r} and RF-Agent \cite{gao2026rf}.
Without module-level tracking, credit assignment, and semantic recombination, these methods struggle to reuse effective reward components, resulting in redundant and inefficient search.

Although promising, module-level reward optimization introduces several unique challenges beyond function-level optimization.
In contrast to optimizing a complete reward program, module-level refinement requires identifying the semantic roles of individual reward components, determining their contribution to policy improvement, and effectively coordinating interactions among different modules.
Moreover, maintaining a reward pool requires a tradeoff between preserving previously discovered effective modules and exploring new reward components, as naive accumulation may introduce redundancy or conflicting incentives. Therefore, designing a module-level reward optimization framework is a more challenging meta-level optimization problem: beyond leveraging holistic RL feedback for reward evaluation, it must attribute performance changes to individual modules and efficiently discover, refine, and reuse effective reward components.

To address these limitations, we propose \textbf{M}odule-\textbf{L}evel \textbf{R}eward \textbf{E}volution \textbf{F}ramework (MLREF). At the core of MLREF is a \emph{module pool}, a persistent repository consisting of multiple reusable reward components, each targeting a specific aspect of the task. This pool evolves across iterations by accumulating successful modules, refining underperforming ones, and reusing proven components for future reward construction. At each iteration, reward functions are assembled as linear combinations of modules drawn from the pool. The framework integrates three mechanisms: (i) \emph{reflection-based refinement} that leverages task analysis and historical feedback to guide module design; (ii) \emph{hybrid weight optimization} that empirically assigns credit to individual modules; and (iii) a \emph{merge strategy with rollback} that consolidates successful modules across parallel samples while recovering from failures. Together, these mechanisms enable stable, empirically grounded pool evolution: each module's design is informed by task analysis and historical feedback, its contribution is credited empirically rather than assumed, and successful modules are consolidated while failures are rolled back, so the reward improves steadily rather than oscillating across iterations.
Table~\ref{tab:comparison} summarizes the key differences between MLREF and prior LLM-based reward design methods.

\begin{table*}[t!]
\centering
\begin{tabular}{lccc}
\toprule
\textbf{Property} & \textbf{EUREKA} & \textbf{RF-Agent} & \textbf{MLREF (Ours)} \\
\midrule
Optimization Unit
    & Function & Function & \textbf{Module Pool} \\
Module Identity
    & -- & Anonymous & \textbf{Named + Spec.} \\
Cross-iter.\ Persistence
    & -- & Partial & \textbf{Persistent Pool} \\
Credit Assignment
    & Holistic & Holistic & \textbf{Per-Module Hybrid} \\
Failure Recovery
    & -- & Implicit & \textbf{Explicit Rollback} \\
Semantic Guidance
    & -- & -- & \textbf{Task + Feedback Refl.} \\
\bottomrule
\end{tabular}
\caption{Comparison of key design properties across LLM-based reward design methods. MLREF is the only framework that treats modules as first-class, persistently managed objects with per-module credit assignment, semantic guidance, and a principled failure recovery mechanism.}
\label{tab:comparison}
\end{table*}

We validate MLREF on 17 tasks from Isaac Gym and Bi-DexHands. Experiments show that MLREF outperforms state-of-the-art LLM-based reward design methods by 25.2\% in locomotion and 6.6\% in manipulation, and ablation studies verify the contributions of individual components. Evolution analysis further reveals that MLREF maintains stable optimization trajectories while baselines suffer from severe performance oscillation, confirming the effectiveness of the rollback mechanism in achieving iterative stability. 

In summary, our contributions are:
\begin{itemize}
    \item We identify the inability of function-level optimization to efficiently reuse modules as a key driver of performance oscillation in LLM-based reward evolution.
    \item We introduce the \emph{module pool} abstraction, which shifts the optimization object from individual reward functions to a persistent repository of reusable modules, and develop three complementary mechanisms that together enable stable, directed pool evolution.
    \item Across 17 locomotion and manipulation tasks, MLREF sets a new state of the art and, more importantly, exhibits substantially more stable optimization dynamics than prior methods, as confirmed by both ablation and evolution analysis.
\end{itemize}

%% file: sections/02-related_work.tex
\section{Related Work}
\label{sec:related}
Before the advent of LLMs, reward design primarily relied on inverse RL \cite{adams2022survey} and preference-based RL \cite{christiano2017deep}; both require substantial human effort and generalize poorly across tasks. Early LLM-based work used LLMs as black-box generators queried per timestep \cite{kwon2023reward,du2023guiding}, which is computationally prohibitive, or generated code-form rewards without iterative refinement \cite{yu2023language}. Later methods introduced iterative refinement guided by human feedback on visualized behaviors \cite{xie2024text2reward,guo2024utilizing,hazra2025revolve}, but still depend on manual intervention, limiting scalability. EUREKA \cite{ma2024eureka} established the first fully automated pipeline for iterative reward optimization. Building on this, R* \cite{li2025r} introduced AST-based crossover with preference-based parameter tuning; CARD \cite{sun2025large} applied heuristic pre-screening before costly RL training; RF-Agent \cite{gao2026rf} adapted MCTS for reward space search; and FORGE \cite{fan2025forging} incorporated chain-of-thought reasoning and systematic crossover. LLM-based reward design has also been explored in game-playing \cite{wu2023read,li2024auto,yifan2025llm}, multi-agent RL \cite{li2025llm,li2025remac, wei2026automated}, and offline imitation learning \cite{sun2025prof}. None of these methods, however, maintain an explicit module pool for systematic reuse or account for LLM-induced variance in evaluation.

In contrast, MLREF is the first framework to maintain a persistent module pool for systematic module-level accumulation, refinement, and reuse throughout the reward evolution process.

%% file: sections/03-preliminaries.tex
\section{Preliminaries}
\label{sec:prelim}

\subsection{Markov Decision Process and Reinforcement Learning}
We consider sequential decision-making problems formalized as Markov Decision Processes (MDPs) \cite{puterman1990markov}, defined by
$(\mathcal{S}, \mathcal{A}, P, R, \gamma, \rho_0)$ with state space $\mathcal{S}$, action space $\mathcal{A}$, transition kernel $P$, reward function $R : \mathcal{S} \times \mathcal{A} \rightarrow \mathbb{R}$, discount factor $\gamma$, and initial state distribution $\rho_0$. A reinforcement learning algorithm seeks a policy $\pi: \mathcal{S}\rightarrow\Delta(\mathcal{A})$ that maximizes the expected discounted return $J(\pi; R) = \mathbb{E}_{\pi, P, \rho_0}\!\big[\sum_{t} \gamma^{t} R(s_t, a_t)\big]$. Because $R$ is the sole task-specific signal in this objective, the choice of reward function entirely determines the behavior that is learned.

\subsection{Reward Design Problem}
Based on the MDP defined above, a reward design problem (RDP) \cite{lewis2010rewards} is characterized by the tuple $\left\langle \mathcal{E}, \mathcal{R}, \mathscr{A}, F \right\rangle$, where $\mathcal{E}=(\mathcal{S},\mathcal{A},P)$ denotes the environment, $\mathcal{R}$ is the space of candidate reward functions, $\mathscr{A}$ is the policy optimization algorithm, and $F: \Pi \rightarrow \mathbb{R}$ is the fitness function that evaluates the task performance of a policy. For each reward function $R\in\mathcal{R}$, let $\pi_R=\mathscr{A}(\mathcal{E},R)$ denote the policy obtained by optimizing $R$ in $\mathcal{E}$. The goal of an RDP is to identify a reward function whose resulting policy maximizes the fitness: $R^*=\arg\max_{R\in\mathcal{R}}F(\pi_R;\mathcal{E})$.

%% file: sections/04-method.tex
\section{Method}
\label{sec:method}

\subsection{Overview}

Inspired by the prior work \cite{ma2024eureka}, MLREF models a reward function as a linear combination of reward modules, where each module targets a specific aspect of the task and, when combined, forms a robust reward function that balances multiple desiderata. Formally, a reward function $R$ is constructed from a set of $K$ modules $\{m_1, m_2, \dots, m_K\}$ with corresponding weights $\{w_1, w_2, \dots, w_K\}$:

\begin{equation}
R(s, a) = \sum_{k=1}^{K} w_k \cdot m_k(s, a).
\label{eq:reward_combination}
\end{equation}

\begin{figure*}[tb]
\centering
\includegraphics[width=0.9\textwidth]{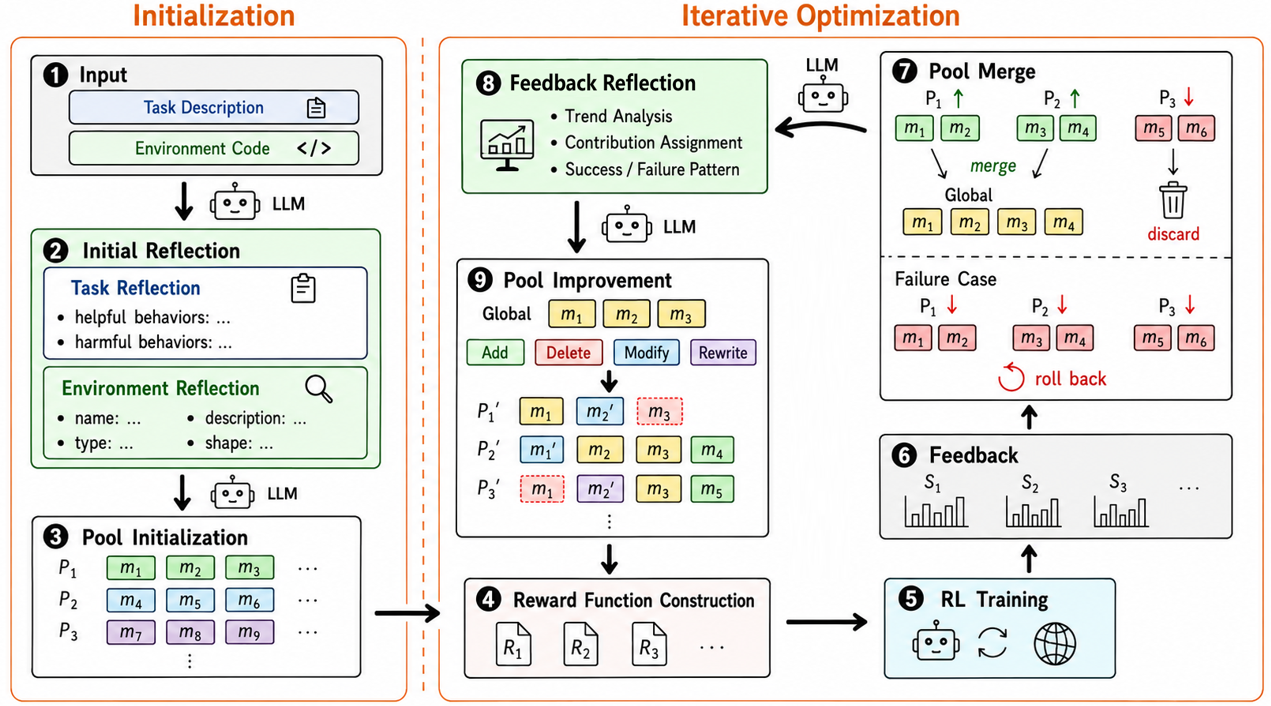}
\caption{Overview of the MLREF pipeline. In the initialization phase (steps~1--3), the LLM performs task and environment reflection and generates diverse module pool variants. In the iterative optimization phase (steps~4--9), each variant assembles a reward function via weighted linear combination for RL evaluation; a merge strategy with rollback consolidates successful modules; and feedback reflection guides the next round of pool improvement.}
\label{fig:pipeline}
\end{figure*}

To fully leverage the potential of these modules, MLREF introduces a \emph{module pool}, a persistent repository where modules are accumulated, refined, and recombined across iterations. Figure~\ref{fig:pipeline} illustrates the full pipeline and Algorithm~\ref{alg:mpdef} presents the corresponding procedure; the pipeline consists of two phases: initialization and iterative optimization. In the initialization phase, the LLM performs initial reflection on the task description and environment code (steps~1--2), then generates $S$ diverse module pool variants (step~3). In the iterative optimization phase, each variant constructs a reward function via weighted linear combination (step~4), which is evaluated through RL training (step~5). A pool merge step consolidates modules from successful variants and rolls back upon failure based on the training feedback (steps~6--7). The LLM then conducts feedback reflection to analyze training outcomes and generates refined pool variants for the next iteration (steps~8--9). Complete LLM prompt templates are provided in the Technical Supplement.

The key departure from prior work lies in what is being optimized. In MLREF, a reward function is only a temporary instance assembled from the pool for RL evaluation; the pool itself is the persistent object that accumulates evidence and improves over iterations. As a result, effective modules persist beyond the function that introduced them and are actively reused in subsequent iterations; training feedback is applied at the module level, enabling finer-grained and more targeted optimization than function-level rewriting.

\begin{algorithm}[tb!]
\caption{MLREF Iterative Optimization}
\label{alg:mpdef}
\begin{algorithmic}
\REQUIRE Task description $T$, environment code $E$, LLM $\mathcal{M}$, RL trainer $\mathcal{A}$, merge strategy $\textsc{Merge}$, max iterations $I$, parallel samples $S$
\ENSURE Best reward function $R^*$
\STATE $\mathcal{P}_{\text{global}} \gets \varnothing$,\; $\mathcal{P}_{\text{prev}} \gets \varnothing$,\; $\mathcal{S}_{\text{best}} \gets \varnothing$,\; $R^* \gets \varnothing$

\FOR{$\text{iter} = 1$ \textbf{to} $I$}
    \STATE \textit{// Stage 1: Module pool generation or improvement}
    \IF{$\text{iter} = 1$}
        \STATE $\text{ref} \gets \mathcal{M}.\textsc{Reflect}(T, E)$
            \textit{// Initial Reflection}
        \FOR{$i = 1$ \textbf{to} $S$ in parallel}
            \STATE $\mathcal{P}_i \gets \mathcal{M}.\textsc{InitPool}(T, E, \text{ref})$
            \textit{// Pool Initialization}
        \ENDFOR
    \ELSE
        \STATE $\text{ref} \gets \mathcal{M}.\textsc{Reflect}(\mathcal{S}_{\text{prev}}, \mathcal{S}_{\text{best}}, \mathcal{P}_{\text{prev}}, \mathcal{P}_{\text{global}})$
            \textit{// Feedback Reflection}
        \FOR{$i = 1$ \textbf{to} $S$ in parallel}
            \STATE $\mathcal{P}_i \gets \mathcal{M}.\textsc{ImprovePool}(T, E, \text{ref}, \mathcal{P}_{\text{global}})$
            \textit{// Pool Improvement}
        \ENDFOR
    \ENDIF

    \STATE \textit{// Stage 2: Reward construction and RL evaluation}
    \FOR{$i = 1$ \textbf{to} $S$ in parallel}
        \STATE $R_i \gets \textsc{ConstructReward}(\mathcal{P}_i)$
            \textit{// Weighted linear combination (Eq.~\ref{eq:reward_combination})}
        \STATE $\mathcal{S}_i \gets \mathcal{A}.\textsc{Train}(E, R_i)$
    \ENDFOR

    \STATE \textit{// Stage 3: Pool merge and best selection}
    \STATE $\mathcal{P}_{\text{global}} \gets \textsc{Merge}(\{\mathcal{P}_i\}, \{\mathcal{S}_i\})$
        \textit{// Accumulate or rollback}
    \STATE $(\mathcal{S}_{\text{prev}}, \mathcal{P}_{\text{prev}}) \gets \operatorname{argmax}_i\; \mathcal{S}_i$
        \textit{// Select the best pool}
    \IF{$\mathcal{S}_{\text{prev}} > \mathcal{S}_{\text{best}}$}
        \STATE $\mathcal{S}_{\text{best}} \gets \mathcal{S}_{\text{prev}}$,\; $R^* \gets \text{corresponding reward of } \mathcal{S}_{\text{prev}}$
    \ENDIF
\ENDFOR
\RETURN $R^*$
\end{algorithmic}
\end{algorithm}

\subsection{Module Pool}

Below we describe the three stages of pool evolution: initialization, improvement, and merge.

\subsubsection{Pool Initialization}

In the first iteration, MLREF constructs a module pool from scratch. The initialization proceeds in two steps: specification generation and module implementation. A module's specification is its metadata, consisting of the module name, input variable names and types, and a natural-language description of its intended function. In the specification generation step, the LLM generates a set of module specifications that collectively cover the diverse requirements of the task, guided by the task description, environment code, and initial reflection. In the implementation step, the LLM generates Python code for each module based on its specification. Separating specification from implementation improves generation quality by letting the LLM focus on one subtask per response.

\subsubsection{Pool Improvement}

From the second iteration onward, MLREF refines the existing module pool. Improvement also follows a two-step structure: improvement plan design and improvement plan execution. In the plan design step, the LLM produces an improvement plan based on the training statistics and feedback reflection. We allow four types of operations on modules:

\begin{enumerate}
    \item \textbf{Add.} Introduce a new module by providing its specification (analogous to initialization).
    \item \textbf{Delete.} Remove a module that is redundant or empirically harmful.
    \item \textbf{Modify.} Revise a module's code while keeping its specification fixed (e.g., fixing incorrect structure or unsuitable output scale).
    \item \textbf{Rewrite.} Replace both the specification and the code when the specification itself is deficient.
\end{enumerate}

For each operation, the LLM also provides a rationale. In the execution step, the LLM implements the concrete code changes relating to new modules, modified modules and rewritten modules according to the plan. This two-step decomposition again improves output quality by separating planning from coding. By operating at the per-module level, MLREF achieves finer-grained control over reward function optimization than prior function-level approaches.

\subsubsection{Pool Merge}

At each iteration, MLREF produces $S$ module pool variants in parallel and consolidates them via a \emph{module-wise merge}. First, variants whose RL performance falls below the historical best by more than a margin $t$ are discarded. All modules from surviving variants are then collected into a single set; when multiple variants contain modules with the same name, the best-performing version is retained. Since all variants derive from the same base pool, this merge consolidates diverse improvement plans without sacrificing quality. If no variant meets the threshold, MLREF rolls back to the previous pool. This rollback mechanism is central to MLREF's iterative stability: it prevents a single poor iteration from destroying the entire optimization trajectory. The failed attempts are recorded and fed into the next round of feedback reflection, encouraging the LLM to explore different strategies.

\subsection{Reflection}

Despite decomposing complex processes into multiple steps, LLM outputs can still suffer from hallucination (producing invalid code that crashes RL training) or insufficient reasoning (yielding suboptimal module designs). To improve output stability and quality, we introduce a reflection mechanism inspired by prior work \cite{li2025remac,sun2025prof} using chain-of-thought prompting. As shown in Fig.~\ref{fig:pipeline} (steps 2 and 8), MLREF employs two types of reflection: initial reflection before the first iteration, and feedback reflection before each subsequent improvement.

\subsubsection{Initial Reflection}

Before the first pool initialization, MLREF conducts initial reflection in two parts. \emph{Task reflection} prompts the LLM to analyze the task description, identifying conditions for success, behaviors likely to help or hinder performance, and non-obvious strategies worth exploring. We explicitly encourage diversity and forbid concrete code generation at this stage. \emph{Environment reflection} extracts relevant observation variables including their names, types, tensor shapes, and descriptions from the environment code, which provides a reference for downstream module design and a validity check against undefined variables. Task and environment reflections are separated to ensure depth in each, and the resulting analysis is reused during all subsequent pool improvement steps.

\subsubsection{Feedback Reflection}

From the second iteration onward, MLREF performs feedback reflection on the previous round's training outcome before planning pool improvements. 

If training completed successfully, the LLM receives the historical best pool (with its module composition and training results) alongside the previous round's pool, improvement plan, and results. It analyzes reward trends per module, compares module design and composition between the two rounds, and attributes performance changes to specific pool differences. When the previous round achieves a new best, the LLM extracts successful design patterns to reinforce; when performance regresses, it diagnoses weaknesses in the improvement plan and proposes new directions.

If training encountered an error, the LLM instead receives the error trace, the previous improvement plan, and the environment code. It locates the error source, analyzes its cause, and suggests concrete measures to avoid similar failures in future iterations.

As with initial reflection, we encourage the LLM to explore diverse hypotheses and forbid it from generating concrete improvement plans or code during this step, preserving a clean separation between analysis and execution.

\subsection{Weight Optimization}

Assigning appropriate weights to modules is critical: different modules vary in importance and output magnitude, so uniform weighting is suboptimal, while relying solely on the LLM to assign weights leads to inconsistency across iterations. MLREF addresses this with a \emph{hybrid weight optimization} strategy that combines LLM-based semantic judgment with empirical evidence from RL training.

Each module maintains two credit scores. \emph{LLM credit} captures the LLM's assessment of a module's relevance: after pool initialization or improvement, the LLM is prompted to select modules and propose initial weights, which are normalized to form LLM credits (with unselected modules receiving zero). \emph{Correlation credit} captures the module's empirical contribution: we compute the Pearson correlation between the module's reward sequence and the performance curve during training, incorporating a time-lag compensation to account for delayed effects of rewards on performance. Both the raw reward--performance correlation and the correlation of their temporal differences are combined into a single correlation score. To prevent credit values from oscillating across iterations, both LLM and correlation credits are smoothed via exponential moving average.

Before combining the two credits, we normalize them to comparable ranges: linear normalization for the non-negative LLM credit, and softmax normalization for the signed correlation credit (amplifying the distinction between positively and negatively correlated modules). The normalized scores are then fused into a composite score via a weighted sum. For module selection, we apply an upper confidence bound (UCB) term over the composite score to balance exploitation of high-scoring modules with exploration of underused ones. The top-$K$ modules by UCB score are selected, and their composite scores are used directly as the weights $w_k$ in the linear combination (Eq.~\ref{eq:reward_combination}). The full mathematical formulation is provided in the Technical Supplement.

%% file: sections/05-experiments.tex
\section{Experiments}
\label{sec:experiments}

\subsection{Experimental Setup}

We evaluate MLREF on 17 representative tasks from Isaac Gym \cite{makoviychuk2021isaacgymhighperformance} and Bi-DexHands \cite{NEURIPS2022_217a2a38}, covering locomotion and dexterous manipulation challenges (see Table~\ref{tab:main_results} for the full list) to answer the following questions: (i) How does MLREF perform compared to state-of-the-art baselines? (ii) How do individual components of MLREF contribute to its performance? (iii) How stable is the reward evolution process under MLREF? 

We compare against two state-of-the-art baselines:

\textbf{EUREKA} \cite{ma2024eureka} generates reward function variants at each iteration by prompting the LLM to improve upon the previous best candidate, using the task description, environment code, and training feedback. We follow the original configuration: 5 iterations with 16 samples per iteration.

\textbf{RF-Agent} \cite{gao2026rf} maintains a Monte Carlo tree over the reward space, selecting promising nodes for targeted optimization, evaluating them via RL, and back-propagating the results. We set the number of MCTS updates to 80, matching the total RL training budget of EUREKA.

For MLREF, to mitigate the randomness introduced by the LLM, we run each task for 3 independent complete pipeline executions and return the reward function from the best execution. Each execution consists of $I=9$ iterations with $S=3$ parallel module pool variants per iteration, yielding a total RL training budget comparable to the baselines. 

For each algorithm on each task, RL training runs for 3,000 epochs within each iteration. After the final reward function is produced, we evaluate it by training a policy from scratch for 20,000 epochs using PPO with 5 independent random seeds, following the RL configuration of EUREKA \cite{ma2024eureka}.

All LLM calls use the DeepSeek-V4-Flash model \cite{xu2026deepseek} in reasoning mode via the official API. Full hyperparameter configurations and LLM prompt templates are provided in the Technical Supplement. The environment setting follows the configuration of EUREKA \cite{ma2024eureka}.

\begin{table*}[!ht]
\centering
\begin{tabular}{lccc}
\toprule
\textbf{Task} & \textbf{EUREKA} & \textbf{RF-Agent} & \textbf{MLREF (Ours)} \\
\midrule
\multicolumn{4}{c}{\textit{Locomotion}} \\
\midrule
Allegro Hand          & 25.48 $\pm$ 1.27  & 27.41 $\pm$ 1.90  & \textbf{28.89 $\pm$ 1.36} \\
Ant                   & 12.35 $\pm$ 1.29  & 12.34 $\pm$ 0.70  & \textbf{12.47 $\pm$ 0.30} \\
Anymal                & \textbf{-0.0057 $\pm$ 0.0021} & -0.0064 $\pm$ 0.0010 & -0.0073 $\pm$ 0.0019 \\
Franka Cabinet        & 0.397 $\pm$ 0.202 & 0.701 $\pm$ 0.200 & \textbf{0.997 $\pm$ 0.006} \\
Humanoid              & \textbf{7.95 $\pm$ 0.87}  & 5.56 $\pm$ 1.63   & 7.40 $\pm$ 1.12 \\
Quadcopter            & -0.042 $\pm$ 0.007  & -0.277 $\pm$ 0.485  & \textbf{-0.040 $\pm$ 0.011} \\
Shadow Hand           & \textbf{17.11 $\pm$ 2.17} & 16.53 $\pm$ 1.65  & 11.75 $\pm$ 9.69 \\
\midrule
Avg.\ Normalized Score & 2.135 & 2.625 & \textbf{3.288}\ (+25.2\%) \\
\midrule
\multicolumn{4}{c}{\textit{Manipulation}} \\
\midrule
Block Stack           & 0.116 $\pm$ 0.050 & 0.139 $\pm$ 0.096 & \textbf{0.313 $\pm$ 0.244} \\
Bottle Cap            & \textbf{0.994 $\pm$ 0.004} & 0.992 $\pm$ 0.010 & 0.987 $\pm$ 0.012 \\
Catch Abreast         & \textbf{0.718 $\pm$ 0.066} & 0.657 $\pm$ 0.047 & 0.644 $\pm$ 0.059 \\
Catch Underarm        & \textbf{0.910 $\pm$ 0.010} & 0.909 $\pm$ 0.016 & 0.871 $\pm$ 0.033 \\
Door Close Outward    & 0.266 $\pm$ 0.129 & \textbf{0.415 $\pm$ 0.265} & 0.256 $\pm$ 0.084 \\
Grasp and Place       & \textbf{0.710 $\pm$ 0.345} & 0.182 $\pm$ 0.130 & 0.475 $\pm$ 0.040 \\
Kettle                & 0.441 $\pm$ 0.376 & \textbf{1.000 $\pm$ 0.000} & 0.901 $\pm$ 0.198 \\
Lift Underarm         & 0.569 $\pm$ 0.465 & 0.512 $\pm$ 0.394 & \textbf{0.930 $\pm$ 0.022} \\
Over                  & 0.968 $\pm$ 0.007 & \textbf{0.985 $\pm$ 0.002} & 0.966 $\pm$ 0.011 \\
Swing Cup             & \textbf{0.945 $\pm$ 0.093} & 0.721 $\pm$ 0.363 & 0.736 $\pm$ 0.294 \\
\midrule
Avg.\ Performance     & 0.664 & 0.651 & \textbf{0.708}\ (+6.6\%) \\
\bottomrule
\end{tabular}
\caption{Performance comparison across 17 tasks. Each row shows the mean and standard deviation of the raw performance metric. The score of locomotion tasks is normalized before averaging. The percentage improvement is relative to the best baseline. Bold indicates the best mean result.}
\label{tab:main_results}
\end{table*}

\subsection{Main Results}

To answer Q1, table~\ref{tab:main_results} reports the performance of MLREF compared to EUREKA and RF-Agent across all 17 tasks. We report the mean and standard deviation of the raw performance metrics. For locomotion tasks, we further report the normalized score $\frac{\text{raw} - \text{sparse}}{\text{human} - \text{sparse}}$, where sparse and human baselines are taken from RF-Agent \cite{gao2026rf}; higher values indicate better performance. Bold indicates the best mean result.

Across the 17 tasks, MLREF achieves the best average performance in both categories. For locomotion, MLREF attains the highest mean on 4 of 7 tasks, with an average normalized score of 3.288, representing a 25.2\% improvement over the best baseline. The gain is most striking on Franka Cabinet, where MLREF approaches the maximum possible score (0.997 vs.\ 0.701 of the nearest baseline). For manipulation, although MLREF obtains the best individual result on only 2 of 10 tasks, its average performance (0.708) is the highest among all methods, yielding a 6.6\% improvement over the best baseline, with particularly strong results on Block Stack and Lift Underarm. These results demonstrate that systematic module-level management via the module pool leads to more effective reward functions than function-level optimization approaches.

\subsection{Ablation Study}

\begin{table*}[t!]
\centering
\begin{tabular}{lcccc}
\toprule
\textbf{Method} & \textbf{Ant} & \textbf{Block Stack} & \textbf{Bottle Cap} & \textbf{Avg.\ \% of Full} \\
\midrule
Full MLREF         & \textbf{12.47 $\pm$ 0.30} & \textbf{0.313 $\pm$ 0.244} & 0.987 $\pm$ 0.012 & \textbf{100.0\%} \\
GPT-4o             & 10.77 $\pm$ 1.95 & 0.101 $\pm$ 0.042 & 0.041 $\pm$ 0.027 & 40.9\% \\
No Pool            & 11.51 $\pm$ 0.64 & 0.085 $\pm$ 0.040 & \textbf{0.988 $\pm$ 0.010} & 73.2\% \\
No Reflection      & 9.58 $\pm$ 0.55 & 0.057 $\pm$ 0.032 & 0.639 $\pm$ 0.401 & 53.3\% \\
No Weight Opt.     & 11.68 $\pm$ 1.14 & 0.044 $\pm$ 0.025 & 0.829 $\pm$ 0.338 & 63.9\% \\
\bottomrule
\end{tabular}
\caption{Ablation results on three representative tasks. Best results are bold.}
\label{tab:ablation}
\end{table*}

To answer Q2, we conduct ablation experiments on three representative tasks (Ant, Bottle Cap, Block Stack) by removing or replacing individual mechanisms from the full MLREF framework:

\begin{enumerate}
    \item \textbf{GPT-4o}: replace the LLM backbone (DeepSeek-V4-Flash) with GPT-4o model \cite{openai2024gpt4o}.
    \item \textbf{No Pool}: remove the module pool; at each iteration, the LLM directly generates a complete reward function, analogous to EUREKA.
    \item \textbf{No Reflection}: remove both initial and feedback reflection; the LLM generates and improves pools directly from the task description and raw training feedback.
    \item \textbf{No Weight Opt.}: remove the hybrid weight optimization strategy; the LLM selects modules and assigns weights without empirical credit assignment.
\end{enumerate}

Table~\ref{tab:ablation} reports the results. All other experimental settings remain identical to the full MLREF configuration.

The full MLREF framework achieves the best or near-best performance across all tasks. Replacing the LLM backbone with GPT-4o causes the most severe degradation (Avg.\ 40.9\%), though MLREF still improves over iterations, indicating that pool evolution can accumulate effective modules even from a weak initialization. Disabling reflection incurs the most consistent degradation (Avg.\ 53.3\%), highlighting its critical role in stabilizing optimization. The No Pool variant matches the full framework on Bottle Cap (0.988 vs.\ 0.987) due to serendipitous exploration, but without the pool, successful designs cannot be stably inherited. We also observed that only the full MLREF framework consistently generalized from validation to test: across all three tasks, the reward selected by its highest validation score remained the best at test time. In contrast, the ablated variants frequently produced rewards that scored well during validation but underperformed at test, suggesting a tendency to overfit when individual components are removed.

\subsection{Iterative Dynamics}

\begin{figure*}[!ht]
\centering
\includegraphics[width=0.33\textwidth]{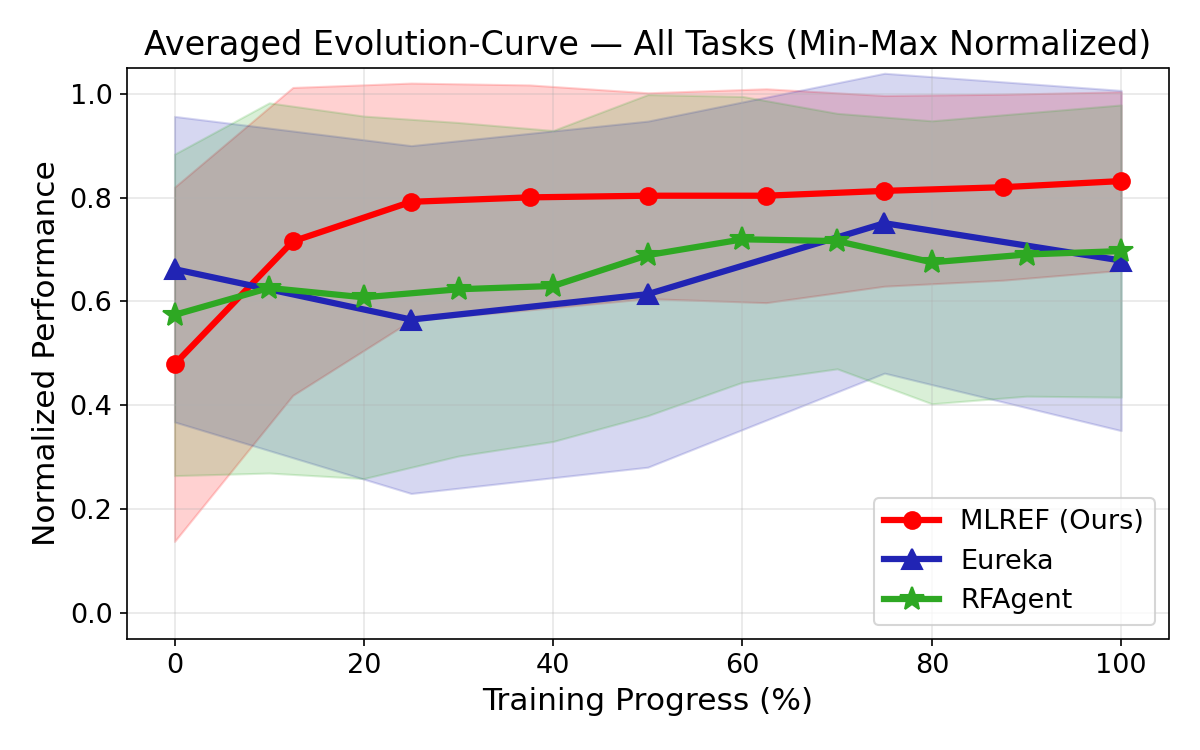}\hfill
\includegraphics[width=0.33\textwidth]{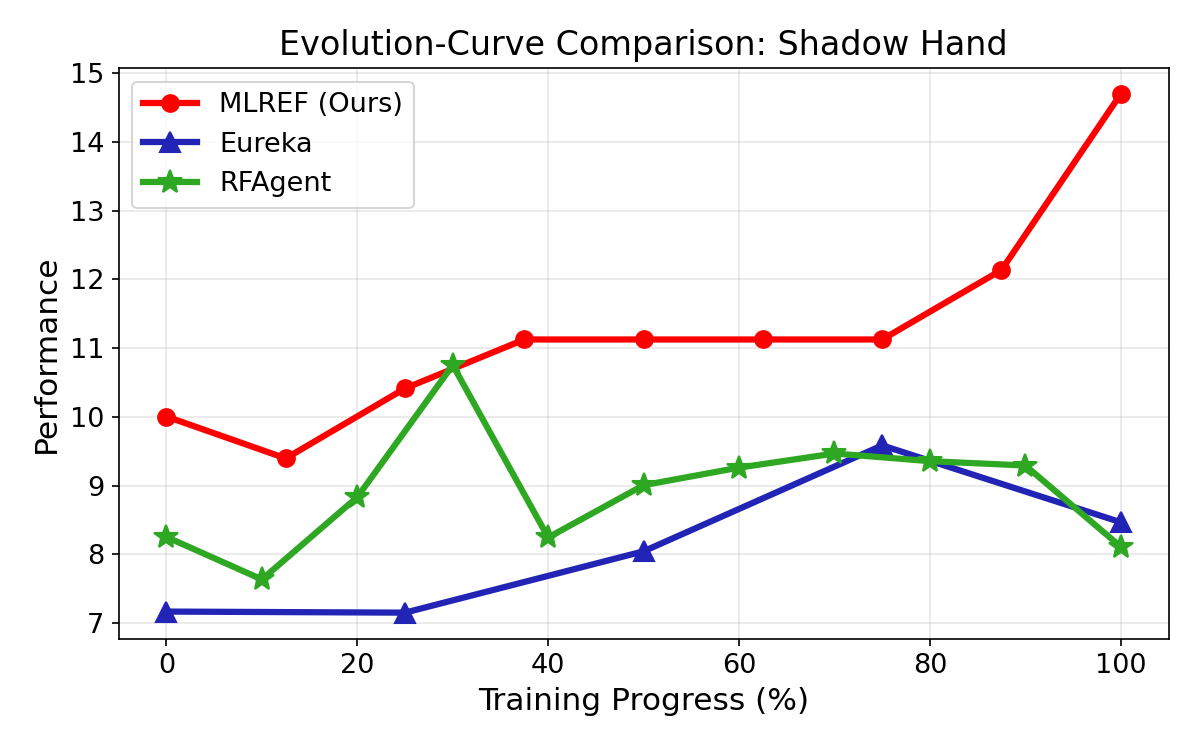}\hfill
\includegraphics[width=0.33\textwidth]{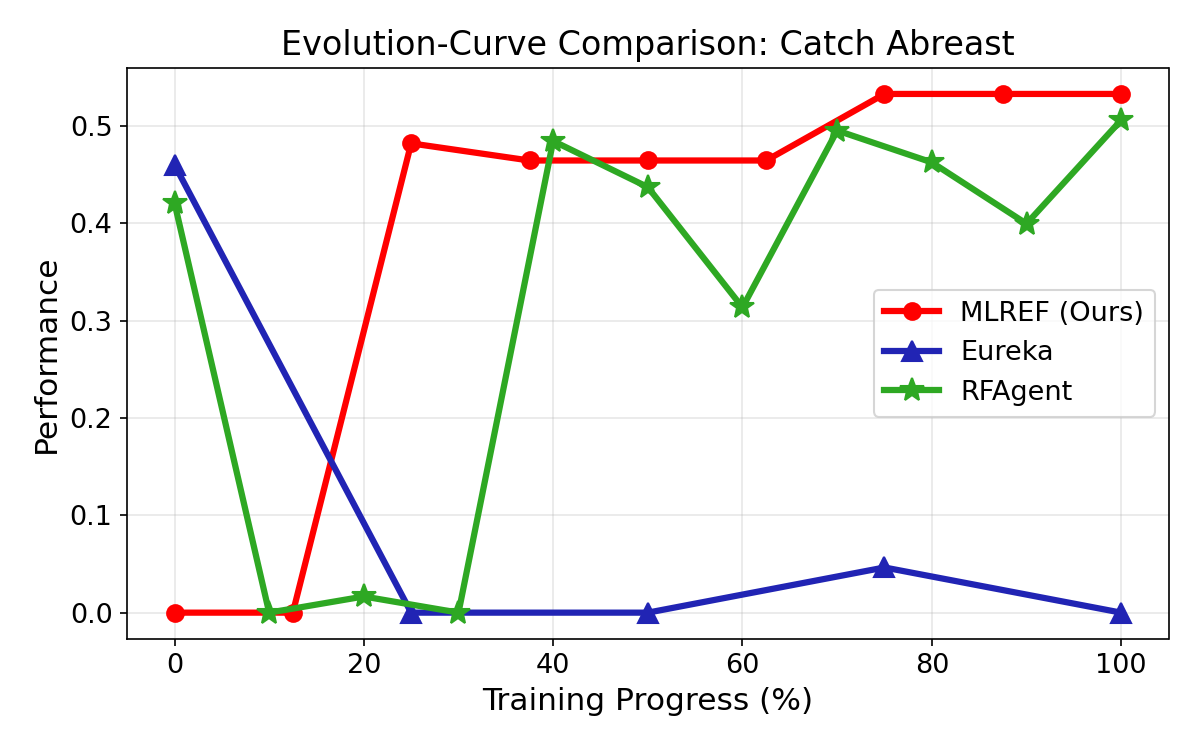}
\caption{Performance evolution across iterations (best sample per iteration). Left: all-task average. Middle: Shadow Hand. Right: Catch Abreast. Shaded areas denote standard deviation.}
\label{fig:evolution}
\end{figure*}

To answer Q3, Figure~\ref{fig:evolution} presents the performance evolution across optimization iterations on average (left panel) and for two representative tasks (middle and right panels). For each algorithm, we track the performance of the best sample at each iteration: for MLREF, this is the best among all parallel pool variants across 3 independent runs; for RF-Agent, the best node expanded at each MCTS step; for EUREKA, the best among the 16 samples of that iteration. Since the three methods operate with different iteration counts, we normalize the horizontal axis to $[0,1]$ according to evolution progress. For the all-task average, we min-max normalize the per-task performance to ensure that each task contributes equally to the aggregate curve.

On the all-task average, EUREKA benefits from large sample counts in early iterations but plateaus thereafter, while MLREF, starting from a lower initial point, exhibits consistent improvement through modular directed optimization; RF-Agent's MCTS-based strategy yields intermediate behavior. On Shadow Hand and Catch Abreast, baselines suffer from severe performance oscillation. In particular, EUREKA degrades sharply after a strong first iteration on Catch Abreast. By contrast, MLREF maintains stable trajectories through its rollback mechanism while continuing to improve via iterative refinement, demonstrating strong optimization stability. Complete evolution curves for all 17 tasks are provided in the Technical Supplement.

\subsection{Discussion}

Our results demonstrate the effectiveness of module-level management. Several limitations suggest directions for future work. First, on sparse-reward manipulation tasks such as Block Stack, MLREF's absolute performance remains modest, suggesting stronger early exploration and progressive exploitation could improve the balance. 
Second, extending the evaluation of MLREF to a broader range of LLMs, including open-source models, would further validate its generality. Finally, while MLREF currently operates with a single LLM, multi-LLM collaboration to leverage diverse reasoning and knowledge represents a promising direction.

%% file: sections/06-conclusion.tex
\section{Conclusion}
\label{sec:conclusion}

We presented MLREF, a module pool-based framework that optimizes reusable reward modules through systematic pool-level operations. Unlike prior function-level approaches, MLREF's module pool enables persistent accumulation, refinement, and reuse of reward components across iterations, supported by reflection, hybrid credit assignment, and a rollback-equipped merge strategy that together achieve iterative stability.

Experiments on 17 tasks demonstrate that MLREF outperforms state-of-the-art LLM-based reward design methods, achieving 25.2\% average improvement in locomotion and 6.6\% in manipulation. Ablation studies confirm the contributions of each component, with reflection being particularly critical. Evolution analysis further shows that MLREF maintains stable optimization trajectories through its rollback mechanism.

%% file: sections/appendix.tex
\section{Weight Optimization Details}

This section provides the mathematical formulation of the hybrid weight optimization strategy discussed in the main paper. We describe the pipeline in three stages: credit assignment, credit normalization and fusion, and module selection.

\subsection{Credit Assignment}

MLREF maintains two credit scores per module: LLM credit and correlation credit.

\subsubsection{LLM Credit}\leavevmode\par

After pool initialization or improvement, the LLM is prompted to select modules and propose initial weights. The assigned weights are normalized to sum to~1, serving as the LLM credit $c_k^{\mathrm{LLM}}$ for the current iteration; unselected modules receive zero credit.

\subsubsection{Correlation Credit}\leavevmode\par

During RL training in the previous iteration, let the performance curve be $P = [p_1, p_2, \dots, p_T]$ and the per-step reward sequence of module $m_k$ be $R = [r_1, r_2, \dots, r_T]$, where $T$ is the number of training steps.

\textit{Smoothing.} To suppress noise, we apply moving-average smoothing with window size $w_s$ to both sequences:

\begin{equation}
P_t^{(s)} = \frac{1}{w_s}\sum_{i=0}^{w_s-1} p_{t+i},\qquad
R_t^{(s)} = \frac{1}{w_s}\sum_{i=0}^{w_s-1} r_{t+i},
\end{equation}

for $t = 1, 2, \dots, T - w_s + 1$.

\textit{Slow difference.} To capture trends, we compute $w_d$-step differences of the smoothed sequences:

\begin{equation}
\Delta P_t = P_{t+w_d}^{(s)} - P_t^{(s)},\qquad
\Delta R_t = R_{t+w_d}^{(s)} - R_t^{(s)},
\end{equation}

for $t = 1, 2, \dots, T - w_s - w_d + 1$.

\textit{Lagged correlation.} To account for delayed effects of rewards on performance, we compute Pearson correlation at multiple lag steps $\ell = 0, 1, \dots, L$:

\begin{equation}
\mathrm{corr}_\ell(X, Y) = \frac{\mathrm{Cov}\big(X_{\ell+1:T'},\; Y_{1:T'-\ell}\big)}{\sigma\big(X_{\ell+1:T'}\big) \cdot \sigma\big(Y_{1:T'-\ell}\big)},
\end{equation}

where $(X, Y)$ can be either the smoothed pair $(P^{(s)}, R^{(s)})$ or the differenced pair $(\Delta P, \Delta R)$, $T'$ is the sequence length after smoothing/differencing, $\mathrm{Cov}(\cdot,\cdot)$ denotes covariance, and $\sigma(\cdot)$ denotes standard deviation.

The raw and differential correlation credits for module $m_k$ are:

\begin{equation}
\begin{aligned}
c_k^{\mathrm{raw}} &= \max_{0 \le \ell \le L} \mathrm{corr}_\ell\big(P^{(s)},\; R^{(s)}\big),\\
c_k^{\mathrm{diff}} &= \max_{0 \le \ell \le L} \mathrm{corr}_\ell\big(\Delta P,\; \Delta R\big).
\end{aligned}
\end{equation}

The final correlation credit is a weighted combination:

\begin{equation}
c_k^{\mathrm{corr}} = w_{\mathrm{raw}} \cdot c_k^{\mathrm{raw}} + (1 - w_{\mathrm{raw}}) \cdot c_k^{\mathrm{diff}},
\end{equation}

where $w_{\mathrm{raw}} \in [0,1]$ balances the contribution of raw versus differential correlation.

\subsubsection{EMA Smoothing}\leavevmode\par

To prevent credit values from oscillating across iterations, both LLM and correlation credits are smoothed via exponential moving average with update rate $\alpha$:

\begin{equation}
\begin{aligned}
\tilde{c}_k^{\,\mathrm{LLM}} &\gets \alpha \cdot c_k^{\mathrm{LLM}} + (1-\alpha) \cdot \tilde{c}_k^{\,\mathrm{LLM}},\\
\tilde{c}_k^{\,\mathrm{corr}} &\gets \alpha \cdot c_k^{\mathrm{corr}} + (1-\alpha) \cdot \tilde{c}_k^{\,\mathrm{corr}}.
\end{aligned}
\end{equation}

Both credits are initialized to~0. Throughout, $c_k$ denotes the raw credit from the current iteration and $\tilde{c}_k$ denotes its exponentially smoothed estimate. A larger $\alpha$ makes the credit more responsive to current observations; a smaller $\alpha$ yields smoother evolution.

\subsection{Credit Normalization and Fusion}

Before fusion, the two smoothed credits $\tilde{c}_i^{\mathrm{LLM}}$ and $\tilde{c}_i^{\mathrm{corr}}$ are normalized to comparable ranges. Let $N$ be the number of modules in the current pool variant.

\textit{LLM credit} is non-negative and normalized linearly:

\begin{equation}
s_i^{\mathrm{llm}} = \frac{\tilde{c}_i^{\mathrm{LLM}}}{\sum_{j=1}^{N} \tilde{c}_j^{\mathrm{LLM}}},
\end{equation}

with a uniform distribution $1/N$ applied if all $\tilde{c}_j^{\mathrm{LLM}} = 0$.

\textit{Correlation credit} $\tilde{c}_i^{\mathrm{corr}} \in [-1, 1]$ is normalized via softmax with temperature $\tau$, subtracting the maximum for numerical stability:

\begin{equation}
s_i^{\mathrm{corr}} = \frac{\exp\!\big((\tilde{c}_i^{\mathrm{corr}} - \max_j(\tilde{c}_j^{\mathrm{corr}})) \,/\, \tau\big)}{\sum_{j=1}^{N} \exp\!\big((\tilde{c}_j^{\mathrm{corr}} - \max_j(\tilde{c}_j^{\mathrm{corr}})) \,/\, \tau\big)}.
\end{equation}

\textit{Fusion.} The normalized scores are fused into a composite score:

\begin{equation}
s_i = w_{\mathrm{LLM}} \cdot s_i^{\mathrm{llm}} + (1 - w_{\mathrm{LLM}}) \cdot s_i^{\mathrm{corr}},
\end{equation}

where $w_{\mathrm{LLM}} \in [0,1]$ controls the relative weight of LLM judgment versus empirical evidence.

\subsection{Module Selection with UCB}

To balance exploitation and exploration, an upper confidence bound (UCB) bonus is added to the composite score:

\begin{equation}
u_i = s_i + \gamma \cdot \frac{\sqrt{\ln(U + 1)}}{u_i + 1},
\end{equation}

where $s_i$ is the fused composite score, $U$ is the total number of times any module from the current pool variant has been selected across iterations, $u_i$ is the selection count of module $m_i$, and $\gamma$ is the UCB exploration coefficient. The top-$K$ modules ranked by $u_i$ are selected, and their composite scores $s_i$ are used directly as the weights $w_i$ in the reward combination (Eq.~1 of the main paper).

\section{Hyperparameters}

Table~\ref{tab:hyperparams} lists all hyperparameters used in MLREF and the baseline methods.

\begin{table*}[t!]
\centering
\begin{tabular}{ll}
\toprule
\textbf{Parameter} & \textbf{Value} \\
\midrule
\multicolumn{2}{c}{\textit{MLREF --- Optimization}} \\
\midrule
Max iterations $I$ & 9 \\
Parallel samples per iteration $S$ & 3 \\
Pipeline runs & 3 \\
RL evaluation seeds & 5 \\
\midrule
\multicolumn{2}{c}{\textit{MLREF --- Pool Merge}} \\
\midrule
Soft update threshold $t$ & 0.1 \\
\midrule
\multicolumn{2}{c}{\textit{MLREF --- Correlation Credit}} \\
\midrule
Smoothing window size $w_s$ & 30 \\
Differential window size $w_d$ & 30 \\
Max lag steps $L$ & 10 \\
Raw correlation weight $w_{\mathrm{raw}}$ & 0.5 \\
\midrule
\multicolumn{2}{c}{\textit{MLREF --- Credit Fusion}} \\
\midrule
EMA update rate $\alpha$ & 0.7 \\
Softmax temperature $\tau$ & 0.25 \\
LLM credit weight $w_{\mathrm{LLM}}$ & 0.5 \\
\midrule
\multicolumn{2}{c}{\textit{MLREF --- Module Selection}} \\
\midrule
UCB exploration rate $\gamma$ & 0.2 \\
Max modules selected $K$ & 5 \\
\midrule
\multicolumn{2}{c}{\textit{RL Training}} \\
\midrule
Algorithm & PPO \\
PPO hyperparameters & Same as EUREKA \cite{ma2024eureka} \\
Environment epochs (iterative optimization) & 3{,}000 \\
Environment epochs (final evaluation) & 20{,}000 \\
\midrule
\multicolumn{2}{c}{\textit{LLM Configuration}} \\
\midrule
Model & DeepSeek-V4-Flash \\
Mode & Reasoning (thinking mode) \\
\midrule
\multicolumn{2}{c}{\textit{Baselines (following original configurations)}} \\
\midrule
EUREKA \cite{ma2024eureka} --- Iterations & 5 \\
EUREKA \cite{ma2024eureka} --- Samples per iteration & 16 \\
RF-Agent \cite{gao2026rf} --- MCTS updates & 80 \\
\bottomrule
\end{tabular}
\caption{Hyperparameters for MLREF and baseline methods.}
\label{tab:hyperparams}
\end{table*}

\section{Iterative Evolution Curves}

This section presents the performance evolution across optimization iterations for all 17 tasks and three domain-level averages. For per-task curve, solid lines denote the best sample for each iteration; for average curves, solid lines and shaded areas represent the mean and standard deviation over tasks. Per-task curves are min-max normalized before averaging to compute domain-level curves.

\begin{figure*}[t!]
\centering
\includegraphics[width=0.23\textwidth]{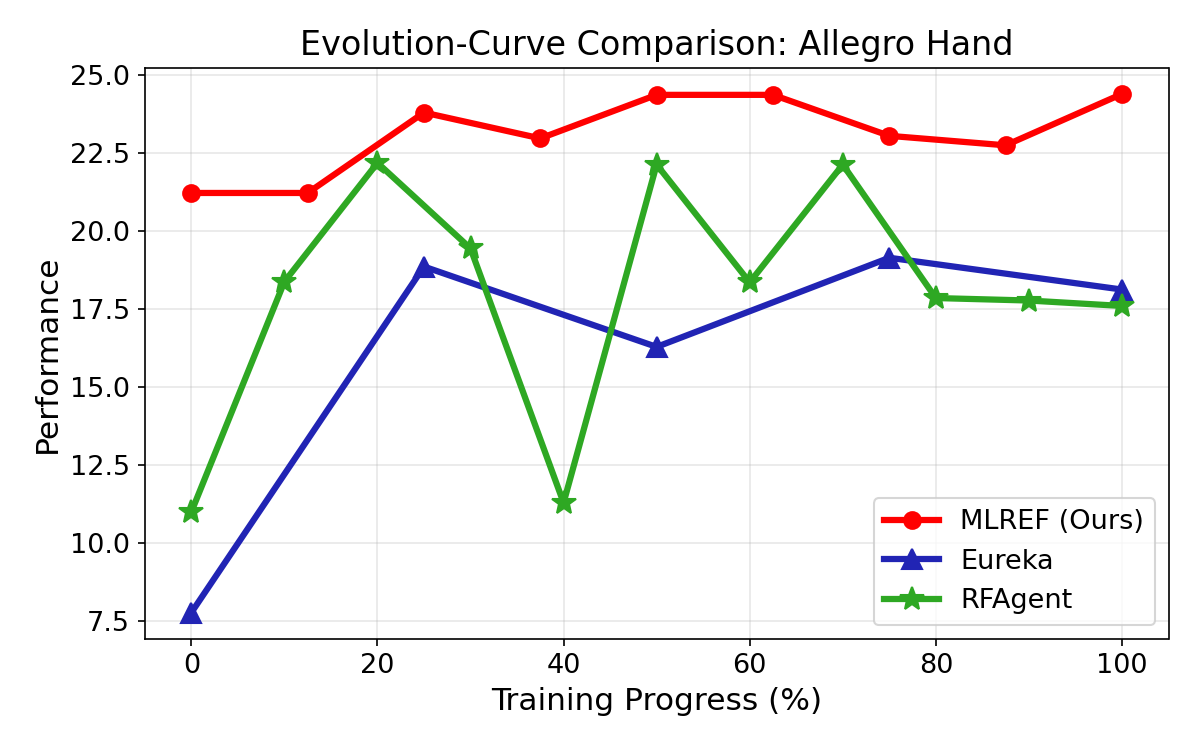}\hfill
\includegraphics[width=0.23\textwidth]{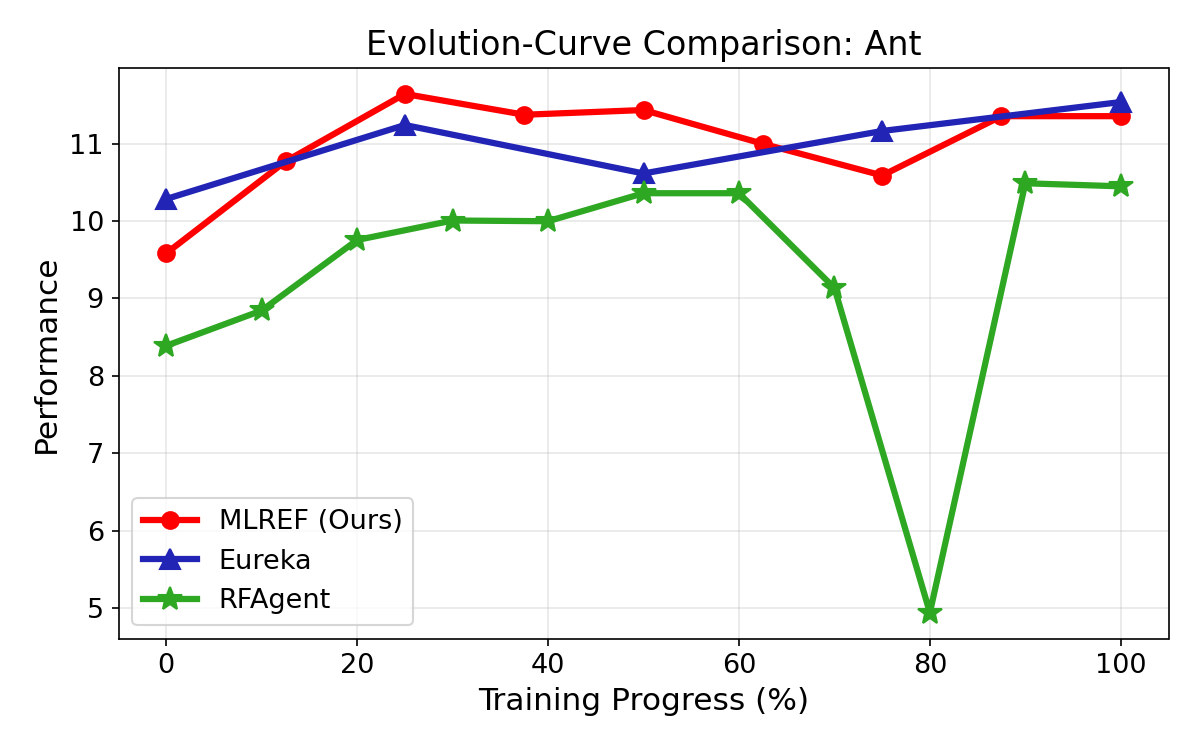}\hfill
\includegraphics[width=0.23\textwidth]{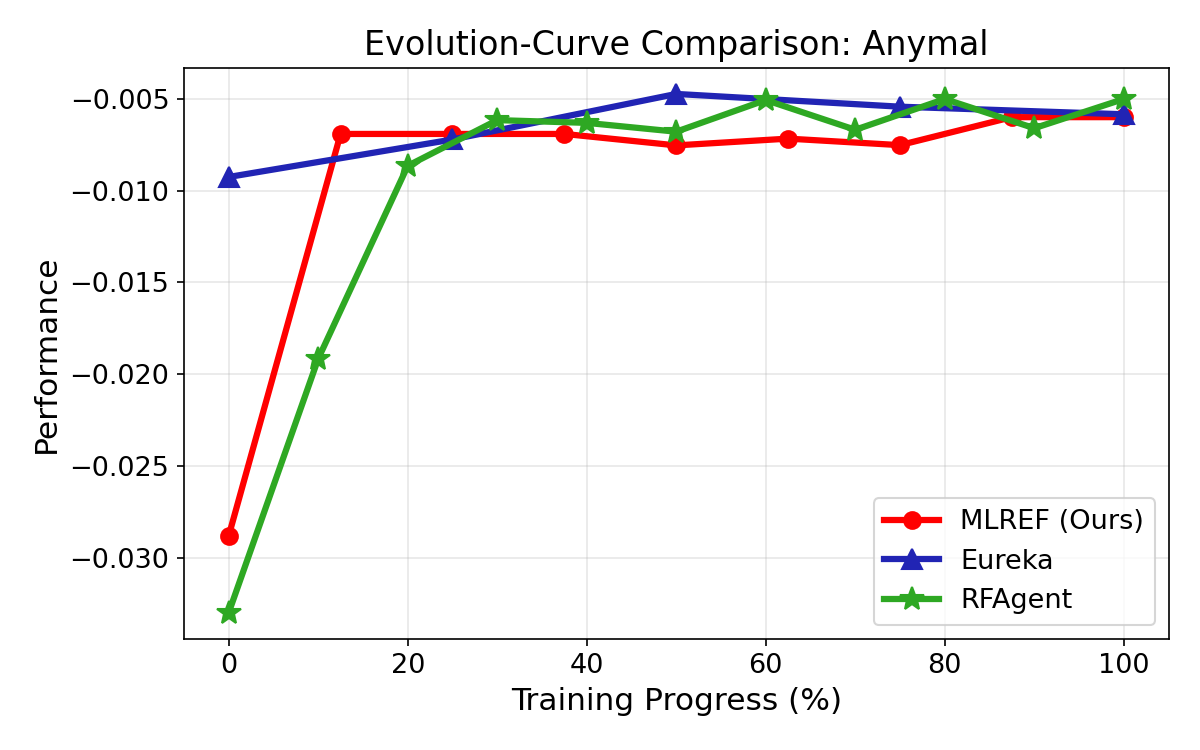}\hfill
\includegraphics[width=0.23\textwidth]{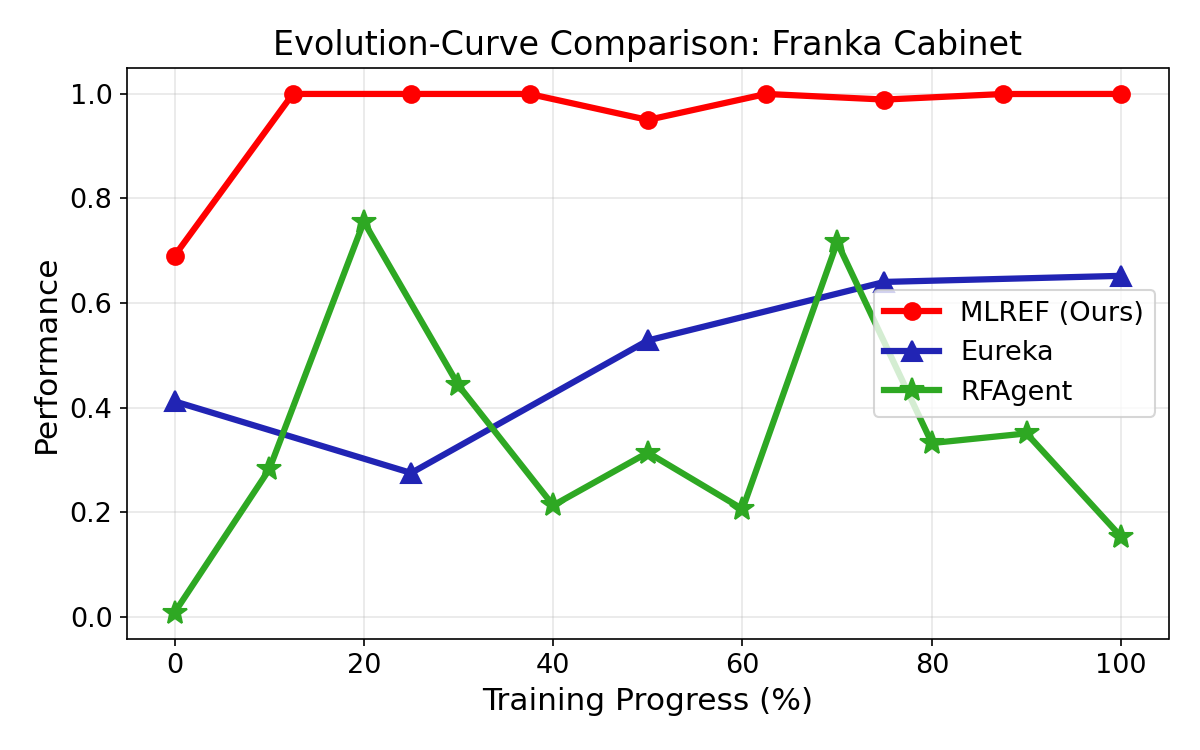}
\\[6pt]
\hfill\includegraphics[width=0.23\textwidth]{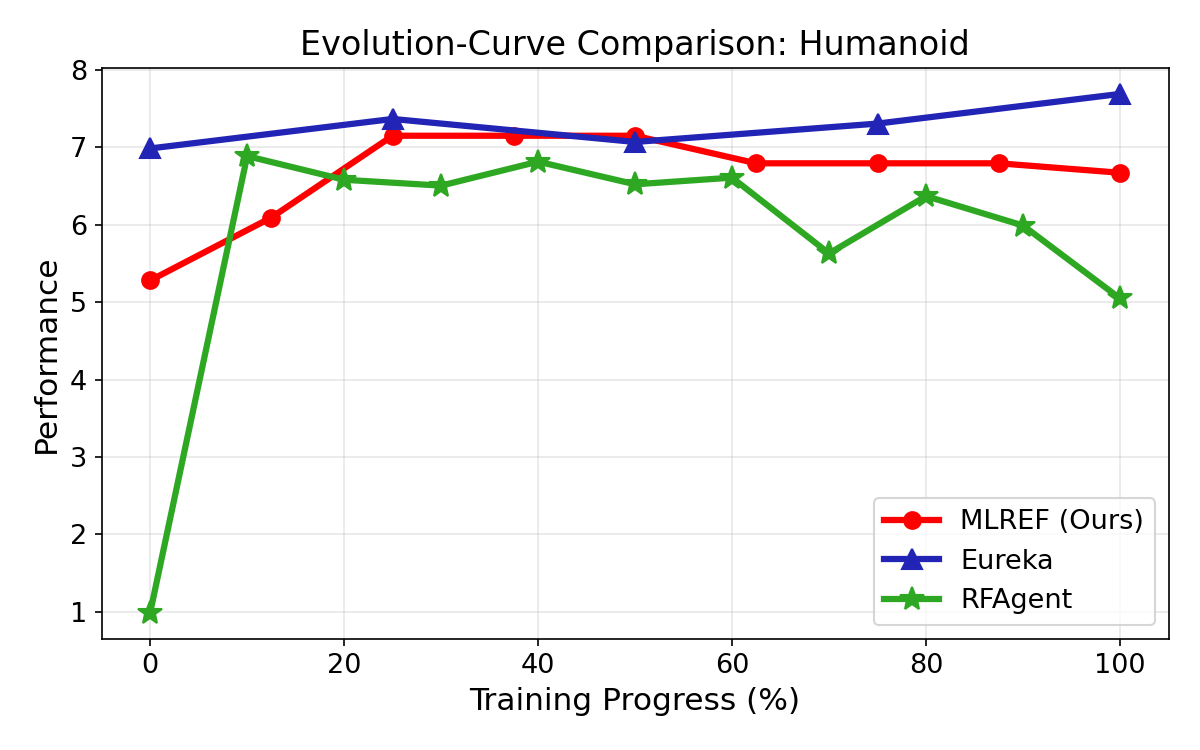}\hfill
\includegraphics[width=0.23\textwidth]{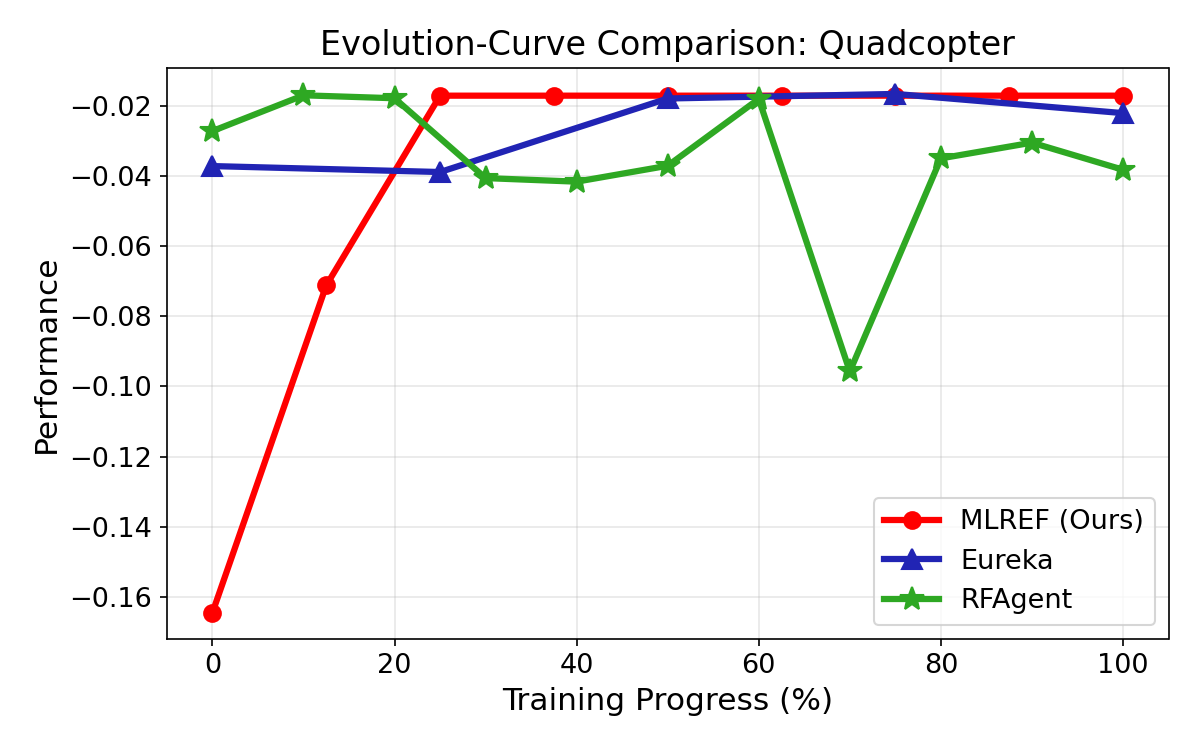}\hfill
\includegraphics[width=0.23\textwidth]{figures/EvolutionCurve/shadow_hand_evolution_comparison.png}\hfill
\caption{Evolution curves for all 7 locomotion tasks.}
\label{fig:supp_evol_loco}
\end{figure*}

\textbf{Locomotion.} Figure~\ref{fig:supp_evol_loco} shows evolution curves for all 7 locomotion tasks.

\begin{figure*}[t!]
\centering
\includegraphics[width=0.23\textwidth]{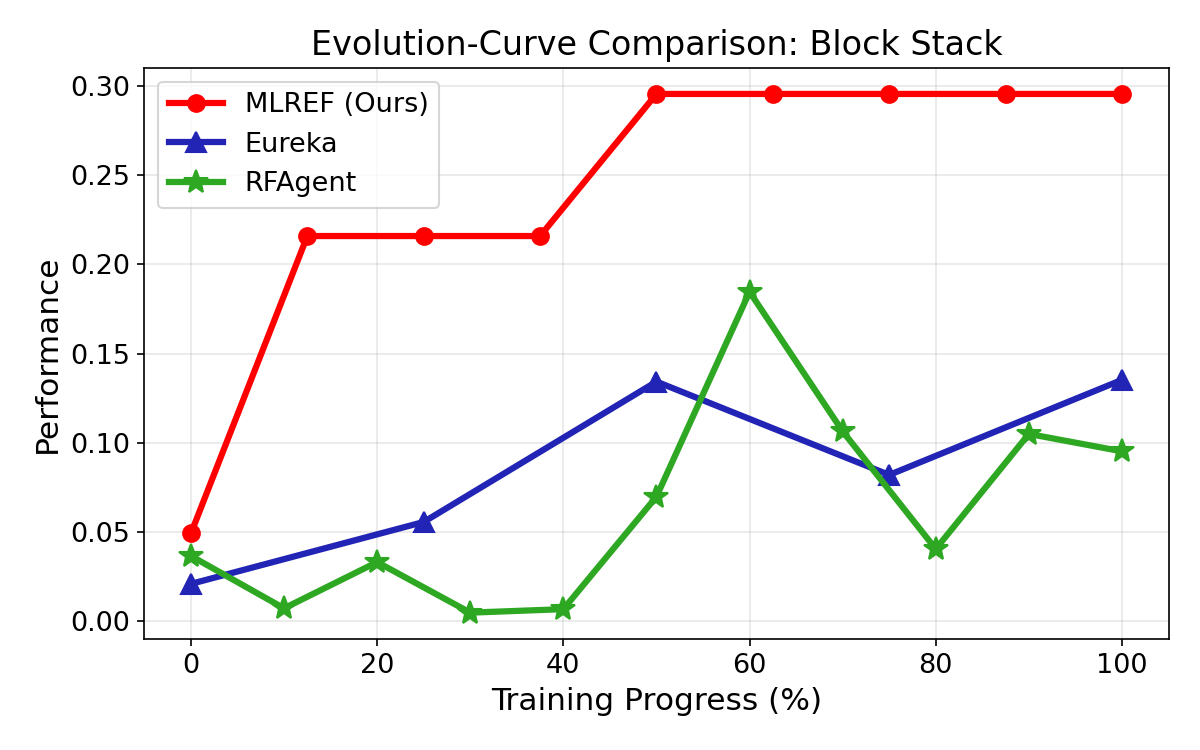}\hfill
\includegraphics[width=0.23\textwidth]{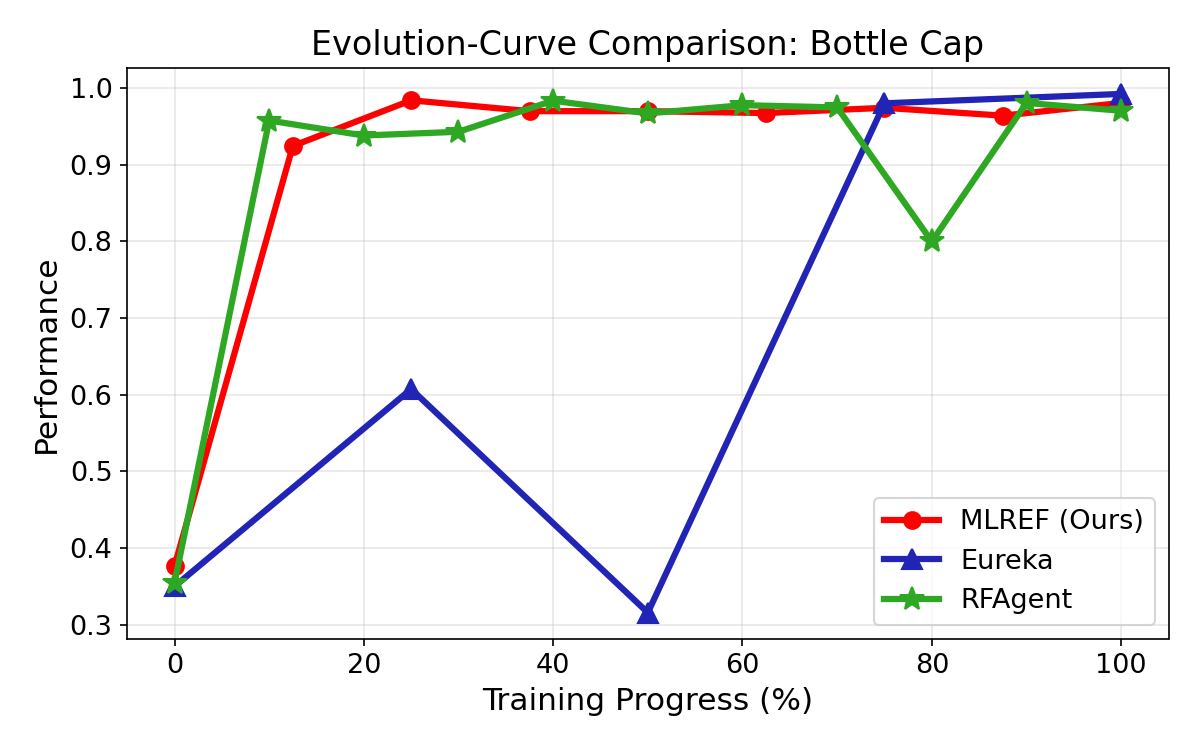}\hfill
\includegraphics[width=0.23\textwidth]{figures/EvolutionCurve/shadow_hand_catch_abreast_evolution_comparison.png}\hfill
\includegraphics[width=0.23\textwidth]{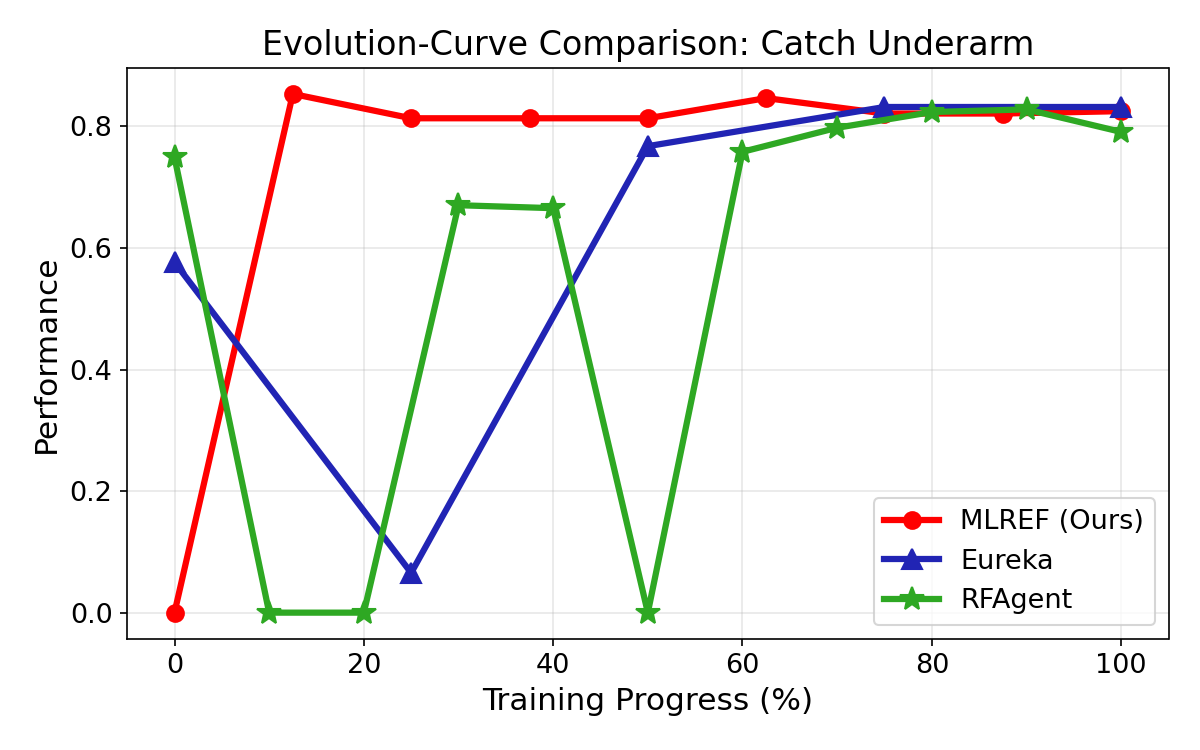}
\\[6pt]
\includegraphics[width=0.23\textwidth]{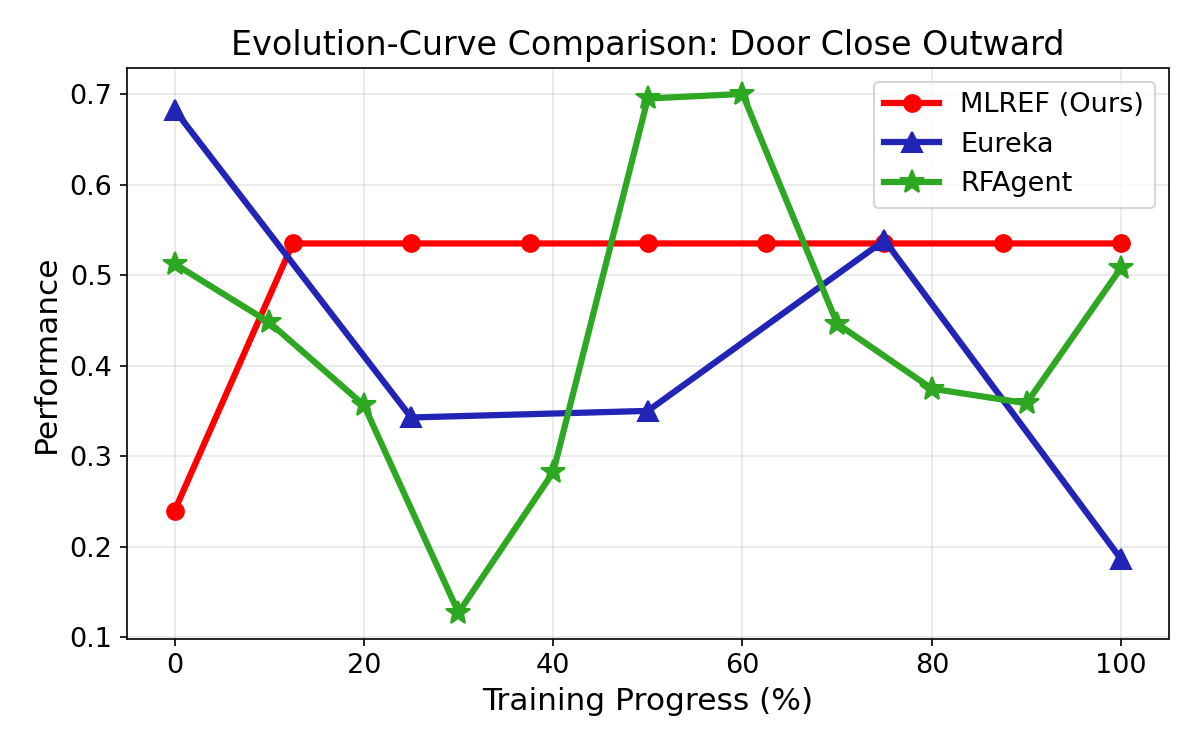}\hfill
\includegraphics[width=0.23\textwidth]{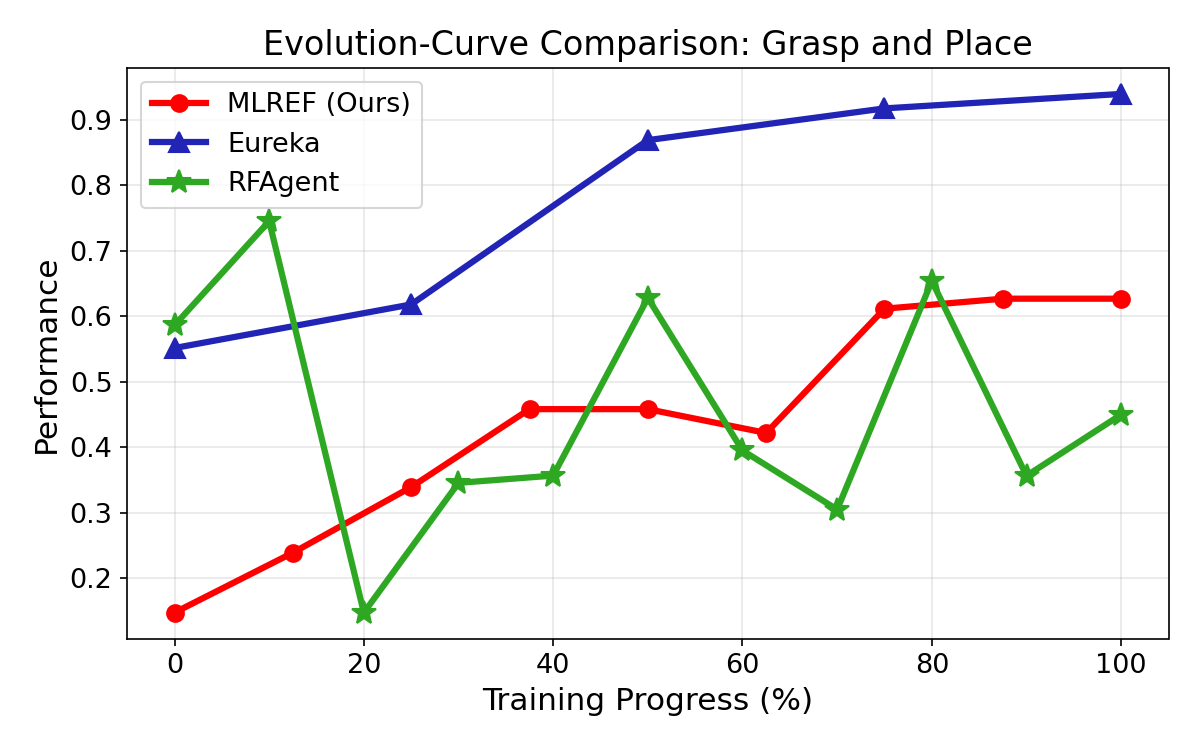}\hfill
\includegraphics[width=0.23\textwidth]{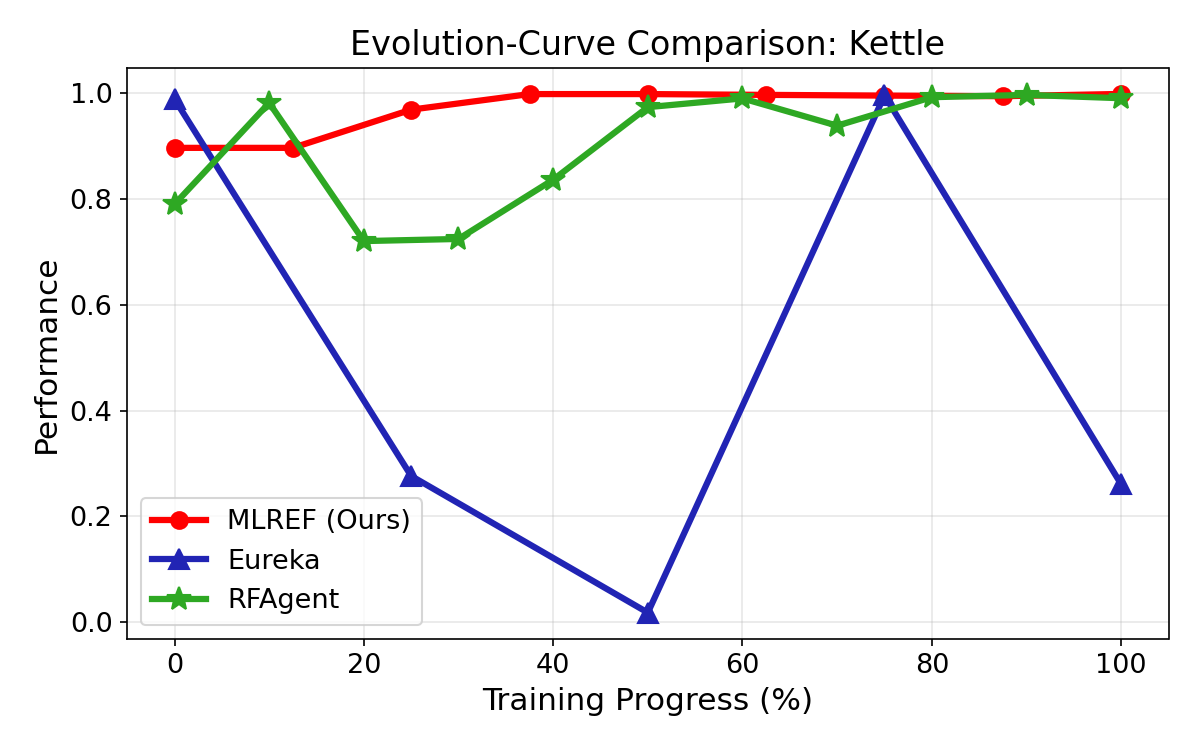}\hfill
\includegraphics[width=0.23\textwidth]{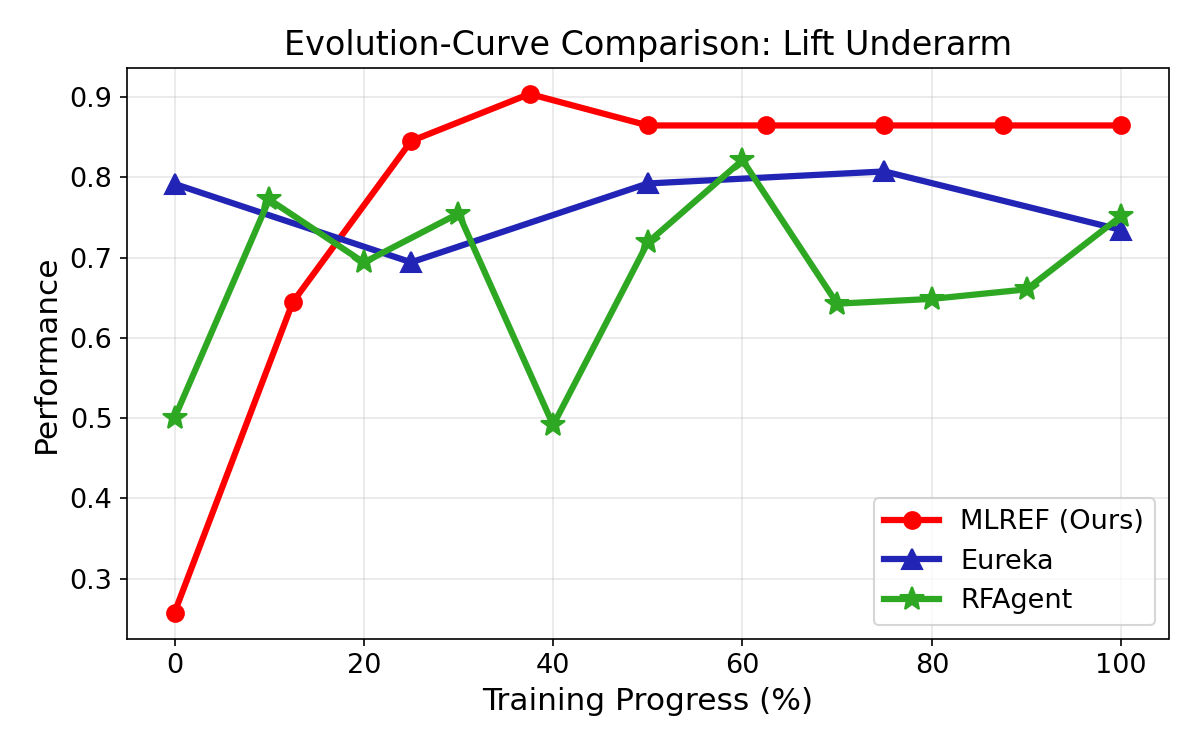}
\\[6pt]
\hfill\includegraphics[width=0.23\textwidth]{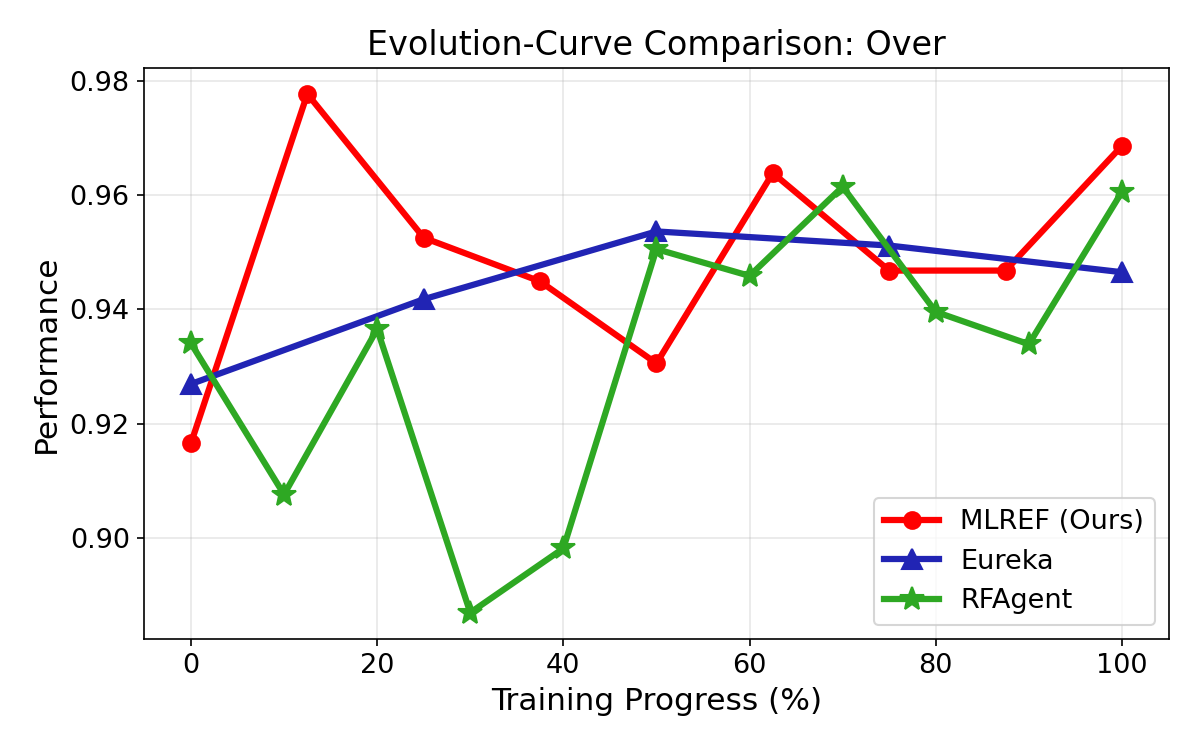}\hfill
\includegraphics[width=0.23\textwidth]{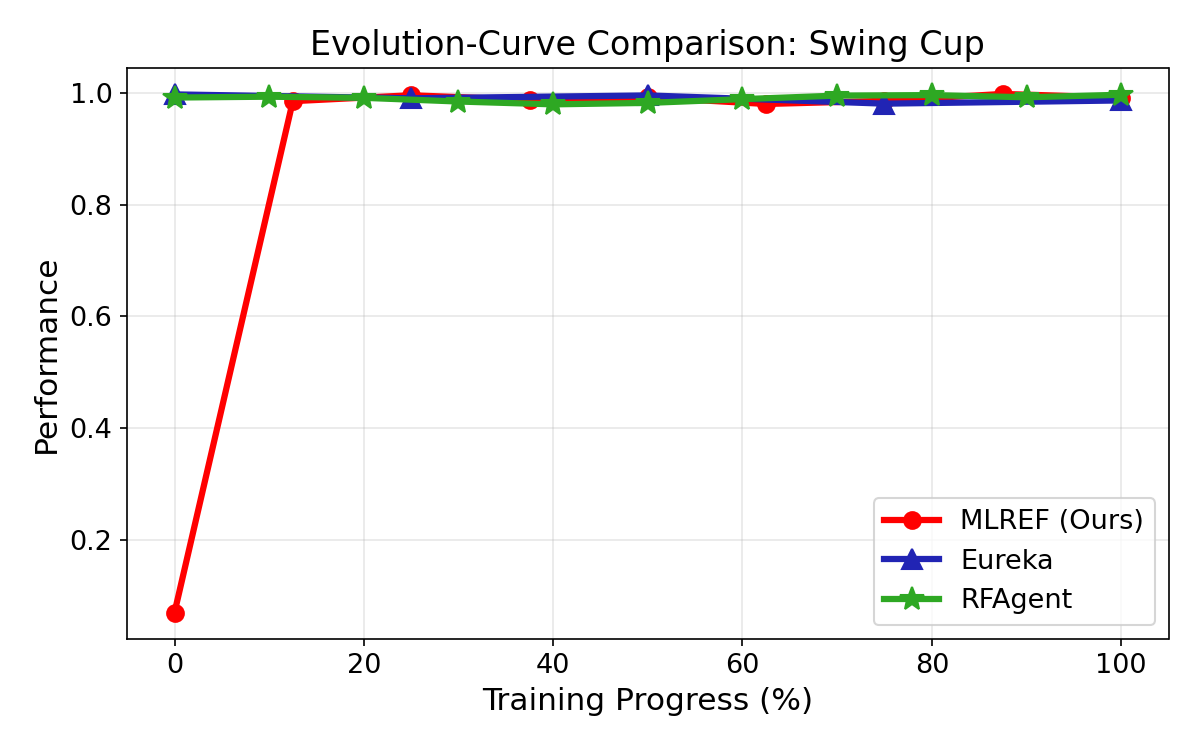}\hfill
\caption{Evolution curves for all 10 manipulation tasks.}
\label{fig:supp_evol_mani}
\end{figure*}

\textbf{Manipulation.} Figure~\ref{fig:supp_evol_mani} shows evolution curves for all 10 manipulation tasks.

\begin{figure*}[t!]
\centering
\includegraphics[width=0.30\textwidth]{figures/EvolutionCurve/averaged_evolution_all_tasks.png}\hfill
\includegraphics[width=0.30\textwidth]{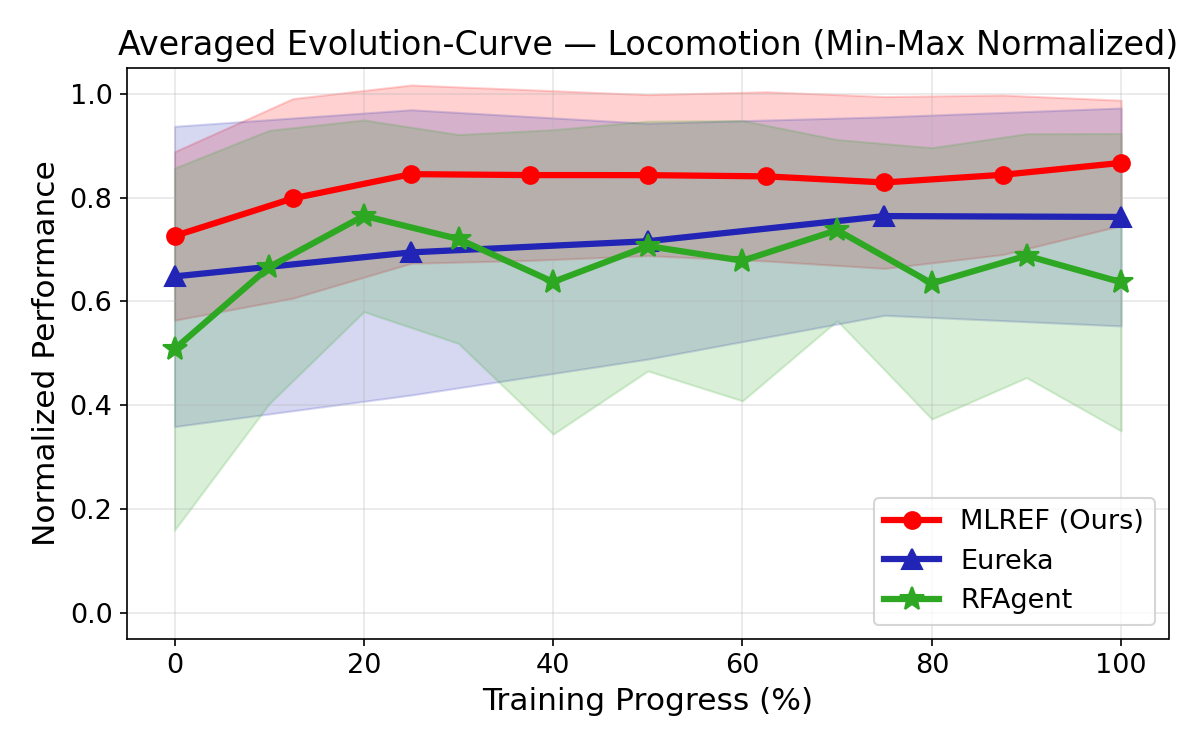}\hfill
\includegraphics[width=0.30\textwidth]{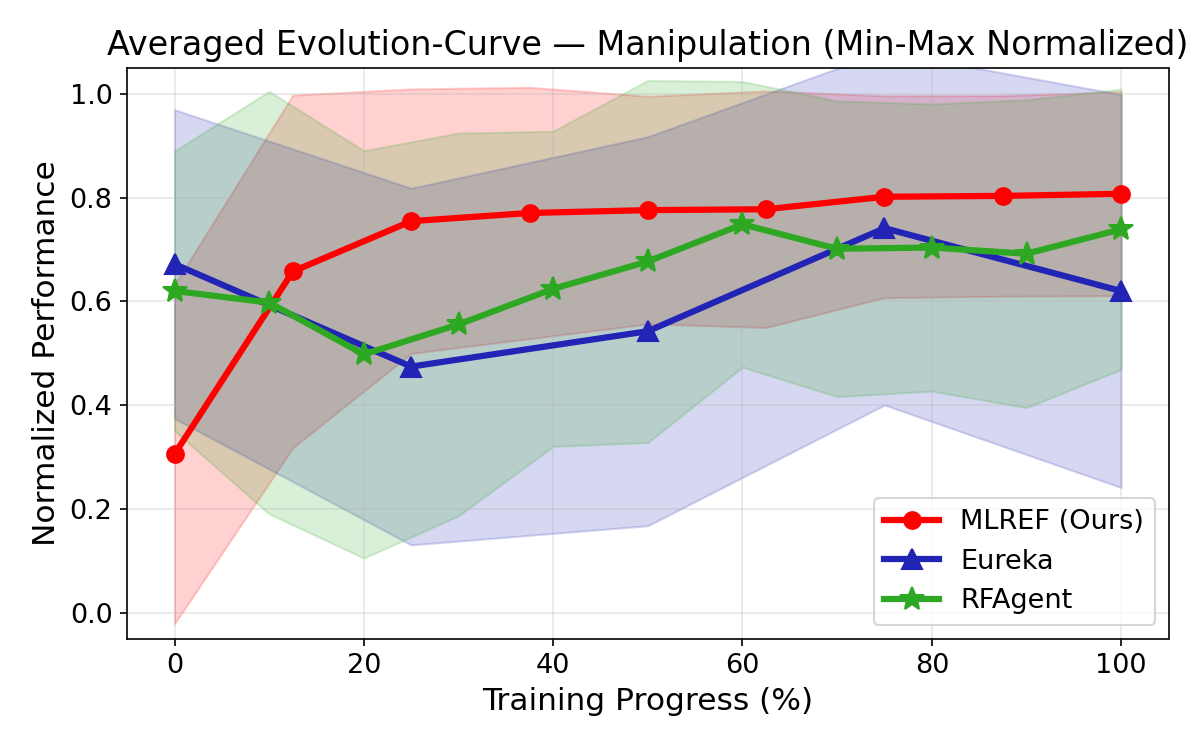}
\caption{Average evolution curves. Left: all 17 tasks. Middle: locomotion (7 tasks). Right: manipulation (10 tasks).}
\label{fig:supp_evol_avg}
\end{figure*}

\textbf{Domain-level averages.} Figure~\ref{fig:supp_evol_avg} reports the average evolution curves.

\section{LLM Prompt Templates}

This section presents the prompts used to invoke the LLM at key stages of MLREF. Content enclosed in curly braces \texttt{\{\}} denotes dynamic fields populated at runtime. Unless otherwise noted, all prompts share the following system-level preamble.

\subsection{System Prompt}

\begin{lstlisting}[breaklines=true, basicstyle=\footnotesize\ttfamily, numbers=none]
You are a reward engineer trying to write reward functions to solve reinforcement learning tasks as effective as possible.
Your goal is to write a reward function for the environment that will help the agent learn the task described in text. 
Your reward function should use useful variables from the environment as inputs. As an example, the reward function signature can be: {task_reward_signature_string}
Since the reward function will be decorated with @torch.jit.script, please make sure that the code is compatible with TorchScript (e.g., use torch tensor instead of numpy array). 
Make sure any new tensor or variable you introduce is on the same device as the input tensors. 

\end{lstlisting}

\subsection{JSON Output Format Constraint}

All prompts that request structured output from the LLM append the following format specification. Individual prompts below omit this block for brevity.

\begin{lstlisting}[breaklines=true, basicstyle=\footnotesize\ttfamily, numbers=none]
You are allowed to reason internally before producing the final answer.
Your final output must be a single JSON object that strictly follows the schema below:
{schema}
Do not include your reasoning or any extra text.

\end{lstlisting}

\subsection{Initial Reflection Prompts}

These prompts are issued once before the first iteration to analyze the task and environment.

\subsubsection{Task Reflection}\leavevmode\par

\begin{lstlisting}[breaklines=true, basicstyle=\footnotesize\ttfamily, numbers=none]
Before designing reward functions, briefly reflect on the task to understand the goal.
Focus on the following:
1. What is the final success condition of the task? Describe only the end result that indicates success.
2. What are some possible behaviors (including non-intuitive or unexpected ones) that could lead to success?
3. What are some common failure patterns where the agent appears to make progress but never actually completes the task?

Important guidelines:
- Do NOT assume the task must be solved in well-defined stages or steps.
- Do NOT assume the behavior needs to be smooth, stable, or human-like.
- Success may arise from aggressive, unstable, or surprising interactions.
- Avoid over-structuring the problem; keep the reasoning flexible and open-ended.

Keep the reflection concise and avoid over-decomposition.
Do NOT design any reward code at this stage. The refined reward code will be designed in the next stage based on your reflections here.

\end{lstlisting}

\subsubsection{Environment Reflection}\leavevmode\par

\begin{lstlisting}[breaklines=true, basicstyle=\footnotesize\ttfamily, numbers=none]
Before designing the reward functions, let's reflect on the following question to better understand the environment:

Question: What are the relevant state variables available in the environment that can be used to design the reward functions? Figure out their types and usages. Also identify their shapes if they are tensors.

Note that the available variables are those defined in the class with prefix `self.` in the environment code.
Do not use the input variables of the functions, as they may be name aliases and not directly accessible in the reward code.

Please include as many variables as possible, since the reward code will be based on these variables.

Do NOT design any reward code at this stage. The refined reward code will be designed in the next stage based on your reflections here.

\end{lstlisting}

\subsection{Pool Initialization Prompts}

These prompts construct the initial module pool from scratch. Generation proceeds in two steps: first specifications, then implementations.

\subsubsection{Specification Generation}\leavevmode\par

\begin{lstlisting}[breaklines=true, basicstyle=\footnotesize\ttfamily, numbers=none]
Design some reward modules that will be included in the module pool.
Just provide the function specifications and a brief description of what each module does.
Each module should may named as "<aspect>_reward" for clarity.
Generate 4 to 6 different reward modules.

Tips for designing modules:
(1) Consider various reward shaping techniques, such as distance-based rewards, progress-based rewards, and task-specific rewards.
(2) Consider make use of every relevant state information available in the environment to design informative reward modules.
(3) Ensure that the modules are diverse and capture different aspects of the task.

\end{lstlisting}

\subsubsection{Module Implementation}\leavevmode\par

\begin{lstlisting}[breaklines=true, basicstyle=\footnotesize\ttfamily, numbers=none]
Implement the python code for a reward module based on this specification:
{specification}

Rules for writing the reward module code:
(1) The module should only contain a single function that computes the reward. Do not include any helper functions.
(2) The return type of the function must be torch.Tensor.
(3) The function name and input variables MUST match the specification. Explicitly specify the type of each input variable and the return type of the function.
(4) The code output should be formatted as a python code string: "```python ... ```".
(5) The code will be run using TorchScript, so it should be compatible with TorchScript.

Some helpful tips for writing the reward function code:
(1) You may find it helpful to normalize the reward to a fixed range by applying transformations like torch.exp to the reward
(2) If you choose to transform a reward component, then you must also introduce a temperature parameter inside the transformation function; this parameter must be a named variable in the reward function and it must not be an input variable. Each transformed reward component should have its own temperature variable

\end{lstlisting}

\subsection{Pool Improvement Prompt}

From the second iteration onward, this prompt refines the existing module pool based on training statistics and feedback reflection.

\begin{lstlisting}[breaklines=true, basicstyle=\footnotesize\ttfamily, numbers=none]
Based on the training statistics of the RL training, please fix the bugs in the existing modules or suggest improvements to the current module pool to better suit the task requirements.
The details of the current module pool are as follows:
{module_pool_details}

You can select from the following types of actions to improve the module pool:
(1) MODIFY: Making small changes to existing modules to better align with the task requirements. You should provide a natural language description of the required changes. The module specification CANNOT be changed in this way.
(2) REWRITE: Making big refactors to existing modules to better align with the task requirements. The module specification CAN be changed in this way, but the name of the module should remain the same for easier tracking.
(3) ADD: Adding new modules to the pool. You should provide a full module specification for the new module without implementing the code.
(4) REMOVE: Removing modules that are redundant or not useful for the task.
For each action, please provide a clear natural language explanation of the reasoning behind it. You can refer to the training statistics if applicable.

Some helpful tips for analyzing the policy feedback:
(1) If the task score is always near zero, then you must consider
    (a) Removing modules that may mislead the agent.
    (b) Adding new modules that better capture important aspects of the task.
(2) If the values for a certain reward component are near identical throughout, then this means RL is not able to optimize this component as it is written. You may consider
    (a) Changing its scale or the value of its temperature parameter.
    (b) Removing it from the pool.
(3) If some reward components' magnitude is significantly larger, then you must re-scale its value to a proper range.
(4) The `compute_reward` function is automatically generated based on the modules and the assembly plan, so you should not directly modify the code of `compute_reward`. Instead, you should modify the modules in the pool and let the system automatically generate the new `compute_reward` function based on the updated modules and assembly plan.

\end{lstlisting}

\subsection{Feedback Reflection Prompts}

The feedback reflection prompt varies depending on the outcome of the previous RL training round. We present the three representative variants below.

\subsubsection{Training Failed}\leavevmode\par

\begin{lstlisting}[breaklines=true, basicstyle=\footnotesize\ttfamily, numbers=none]
The refined reward function in the CURRENT iteration did not run successfully, so we roll back to the previous BEST module pool design as the basis of our reflection and improvement.

Here is the previous BEST module pool for your reference:
{module_pool_details}

Here is the module pool update plan and the error message in the CURRENT iteration:
{improve_plan}
{training_signal}

{env_plugin}

Before designing the new reward functions based on the BEST module pool again, let's reflect on the following questions to better understand the error and identify potential issues in the reward function design:
1. What is the critical error message in the training signal?
2. Where is the error occurring in the code?
3. Based on the provided environment, what could be the potential reasons for this error?
4. How can we avoid this error in the next iteration of module pool improvement?

Please think step by step and provide your reflections on these questions.
Just answer the questions one by one using natural language, and feel free to provide any additional insights or observations that may be relevant.
Do NOT design any reward code at this stage. The refined reward function will be designed in the next stage based on your reflections here.
\end{lstlisting}

\subsubsection{Training Succeeded but Did Not Surpass Best}\leavevmode\par

\begin{lstlisting}[breaklines=true, basicstyle=\footnotesize\ttfamily, numbers=none]
The performance of the CURRENT module pool design has regressed compared to the previously BEST performance, so we roll back to the BEST module pool as the basis of our reflection and improvement.

Here is the BEST module pool for your reference:
{module_pool_details}

Here is the BEST module assembly along with the training statistics from the BEST iteration:
{best_module_usage_list}
{best_training_signal}

Here is the module pool update plan in the CURRENT iteration:
{improve_plan}

Here is the module assembly and statistics from the CURRENT iteration for comparison:
{module_usage_list}
{training_signal}

Before designing new improvement plan based on the BEST module pool again, let's reflect on the following questions to better understand the statistics and identify potential issues in the current module pool design:
1. What are the key trends and overall performance observed in the training statistics?
2. What are the differences in the module design and assembly between the current and best iterations? Are they responsible for the observed performance regression?
3. How does each reward module contribute to the overall performance? Are there any modules that seem to be more effective or less effective based on the statistics? Provide some suggestions on the future use of these modules.
4. Are there any potential issues or limitations in the current reward function design? How can we address these issues in the next iteration?

Please think step by step and provide your reflections on these questions.
Just answer the questions one by one using natural language, and feel free to provide any additional insights or observations that may be relevant.

Notice that we have rolled back to the BEST module pool, so the next step improvement plan will be based on the BEST module pool.
Try to come up with some new ideas to avoid the repeated regression.
Do NOT generate any specific plan at this stage. The pool improvement plan will be designed in the next stage based on your reflections here.

\end{lstlisting}

\subsubsection{Training Succeeded and Surpassed Best}\leavevmode\par

\begin{lstlisting}[breaklines=true, basicstyle=\footnotesize\ttfamily, numbers=none]
Here is the current module pool details for your reference:
{module_pool_details}

Here is the best module assembly along with the training statistics from the CURRENT iteration:
{module_usage_list}
{training_signal}

Here is the best module assembly and statistics from the PREVIOUS iteration for comparison:
{module_usage_list_prev}
{training_signal_prev}

Here is the module pool update plan in the PREVIOUS iteration:
{improve_plan}

Before designing new improvement plan for the module pool, let's reflect on the following questions to better understand the statistics and identify potential issues in the current module pool design:
1. What are the key trends and overall performance observed in the training statistics? Are there any significant improvements or regressions compared to the previous iteration?
2. What are the differences in the module design and assembly between the current and previous iterations? Are they responsible for the performance regression?
3. How does each reward module contribute to the overall performance? Are there any modules that seem to be more effective or less effective based on the statistics? Provide some suggestions on the future use of these modules.
4. Are there any potential issues or limitations in the current reward function design? How can we address these issues in the next iteration?

Please think step by step and provide your reflections on these questions.
Just answer the questions one by one using natural language, and feel free to provide any additional insights or observations that may be relevant.
Do NOT generate any specific plan at this stage. The pool improvement plan will be designed in the next stage based on your reflections here.

\end{lstlisting}

\subsection{Weight Selection Prompt}

This prompt asks the LLM to select modules and assign weights, producing the LLM credit used in hybrid weight optimization.

\begin{lstlisting}[breaklines=true, basicstyle=\footnotesize\ttfamily, numbers=none]
Here is the modified module pool, along with the specifications of each module.
{module_pool}

You should construct the reward function as a linear combination of the modules in the modified pool.
Please choose the most appropriate modules and decide their weights to construct the reward function for the given task.
You can also refer to the training statistics of the previous RL training and the improvement plan to help you find the most promising modules.

Tips for selecting modules:
Consider including these modules in the final reward function:
(1) The modules whose reward values changed a lot during training, as they are likely to be more effective for RL optimization.
(2) The modules that were modified in the improvement plan, as they are expected to better align with the task requirements.
(3) The newly added modules in the improvement plan, as they may capture important aspects of the task that were previously missing.

Tips for choosing weights:
(1) The weights should be non-negative, as all modules are designed to provide positive feedback for desirable behaviors.
(2) The weights should have balanced magnitudes to ensure that no single module dominates the reward signal.
(3) Consider the relative importance of each module in achieving the task objectives when assigning weights.

\end{lstlisting}

%% file: references.bib
@inproceedings{lewis2010rewards,
  title={Where do rewards come from},
  author={Lewis, Richard L and Singh, Satinder and Barto, Andrew G},
  booktitle={Proceedings of the international symposium on AI-inspired biology},
  pages={2601--2606},
  year={2010}
}

@article{puterman1990markov,
  title={Markov decision processes},
  author={Puterman, Martin L},
  journal={Handbooks in operations research and management science},
  volume={2},
  pages={331--434},
  year={1990},
  publisher={Elsevier}
}

@article{radosavovic2024real,
  title={Real-world humanoid locomotion with reinforcement learning},
  author={Radosavovic, Ilija and Xiao, Tete and Zhang, Bike and Darrell, Trevor and Malik, Jitendra and Sreenath, Koushil},
  journal={Science Robotics},
  volume={9},
  number={89},
  pages={eadi9579},
  year={2024},
  publisher={American Association for the Advancement of Science}
}

@article{elguea2023review,
  title={A review on reinforcement learning for contact-rich robotic manipulation tasks},
  author={Elguea-Aguinaco, {\'I}{\~n}igo and Serrano-Mu{\~n}oz, Antonio and Chrysostomou, Dimitrios and Inziarte-Hidalgo, Ibai and B{\o}gh, Simon and Arana-Arexolaleiba, Nestor},
  journal={Robotics and Computer-Integrated Manufacturing},
  volume={81},
  pages={102517},
  year={2023},
  publisher={Elsevier}
}

@article{zhu2021deep,
  title={Deep reinforcement learning based mobile robot navigation: A review},
  author={Zhu, Kai and Zhang, Tao},
  journal={Tsinghua Science and Technology},
  volume={26},
  number={5},
  pages={674--691},
  year={2021},
  publisher={TUP}
}

@inproceedings{goldwaser2020deep,
  title={Deep reinforcement learning for general game playing},
  author={Goldwaser, Adrian and Thielscher, Michael},
  booktitle={Proceedings of the AAAI conference on artificial intelligence},
  volume={34},
  pages={1701--1708},
  year={2020}
}

@article{wei2022chain,
  title={Chain-of-thought prompting elicits reasoning in large language models},
  author={Wei, Jason and Wang, Xuezhi and Schuurmans, Dale and Bosma, Maarten and Xia, Fei and Chi, Ed and Le, Quoc V and Zhou, Denny and others},
  journal={Advances in neural information processing systems},
  volume={35},
  pages={24824--24837},
  year={2022}
}

@article{zhang2024chain,
  title={Chain of preference optimization: Improving chain-of-thought reasoning in llms},
  author={Zhang, Xuan and Du, Chao and Pang, Tianyu and Liu, Qian and Gao, Wei and Lin, Min},
  journal={Advances in Neural Information Processing Systems},
  volume={37},
  pages={333--356},
  year={2024}
}

@article{ouyang2022training,
  title={Training language models to follow instructions with human feedback},
  author={Ouyang, Long and Wu, Jeffrey and Jiang, Xu and Almeida, Diogo and Wainwright, Carroll and Mishkin, Pamela and Zhang, Chong and Agarwal, Sandhini and Slama, Katarina and Ray, Alex and others},
  journal={Advances in neural information processing systems},
  volume={35},
  pages={27730--27744},
  year={2022}
}

@article{chen2021evaluating,
  title={Evaluating large language models trained on code},
  author={Chen, Mark and Tworek, Jerry and Jun, Heewoo and Yuan, Qiming and Pinto, Henrique Ponde De Oliveira and Kaplan, Jared and Edwards, Harri and Burda, Yuri and Joseph, Nicholas and Brockman, Greg and others},
  journal={arXiv preprint arXiv:2107.03374},
  year={2021}
}

@article{adams2022survey,
  title={A survey of inverse reinforcement learning},
  author={Adams, Stephen and Cody, Tyler and Beling, Peter A},
  journal={Artificial Intelligence Review},
  volume={55},
  number={6},
  pages={4307--4346},
  year={2022},
  publisher={Springer}
}

@article{christiano2017deep,
  title={Deep reinforcement learning from human preferences},
  author={Christiano, Paul F and Leike, Jan and Brown, Tom and Martic, Miljan and Legg, Shane and Amodei, Dario},
  journal={Advances in neural information processing systems},
  volume={30},
  year={2017}
}

@article{kwon2023reward,
  title={Reward design with language models},
  author={Kwon, Minae and Xie, Sang Michael and Bullard, Kalesha and Sadigh, Dorsa},
  journal={arXiv preprint arXiv:2303.00001},
  year={2023}
}

@inproceedings{du2023guiding,
  title={Guiding pretraining in reinforcement learning with large language models},
  author={Du, Yuqing and Watkins, Olivia and Wang, Zihan and Colas, C{\'e}dric and Darrell, Trevor and Abbeel, Pieter and Gupta, Abhishek and Andreas, Jacob},
  booktitle={International Conference on Machine Learning},
  pages={8657--8677},
  year={2023},
  organization={PMLR}
}

@article{yu2023language,
  title={Language to rewards for robotic skill synthesis},
  author={Yu, Wenhao and Gileadi, Nimrod and Fu, Chuyuan and Kirmani, Sean and Lee, Kuang-Huei and Arenas, Montse Gonzalez and Chiang, Hao-Tien Lewis and Erez, Tom and Hasenclever, Leonard and Humplik, Jan and others},
  journal={arXiv preprint arXiv:2306.08647},
  year={2023}
}

@inproceedings{xie2024text2reward,
  title={Text2reward: Reward shaping with language models for reinforcement learning},
  author={Xie, Tianbao and Zhao, Siheng and Wu, Chen and Liu, Yitao and Luo, Qian and Zhong, Victor and Yang, Yanchao and Yu, Tao},
  booktitle={International Conference on Learning Representations},
  volume={2024},
  pages={35663--35699},
  year={2024}
}

@inproceedings{hazra2025revolve,
  title={REvolve: Reward evolution with large language models using human feedback},
  author={Hazra, Rishi and Sygkounas, Alkis and Persson, Andreas and Loutfi, Amy and Zuidberg Dos Martires, Pedro},
  booktitle={International Conference on Learning Representations},
  volume={2025},
  pages={101949--101990},
  year={2025}
}

@inproceedings{guo2024utilizing,
  title={Utilizing large language models for robot skill reward shaping in reinforcement learning},
  author={Guo, Qi and Liu, Xing and Hui, Jianjiang and Liu, Zhengxiong and Huang, Panfeng},
  booktitle={International Conference on Intelligent Robotics and Applications},
  pages={3--17},
  year={2024},
  organization={Springer}
}

@inproceedings{ma2024eureka,
  title={Eureka: Human-level reward design via coding large language models},
  author={Ma, Yecheng Jason and Liang, William and Wang, Guanzhi and Huang, De-An and Bastani, Osbert and Jayaraman, Dinesh and Zhu, Yuke and Fan, Jim and others},
  booktitle={International conference on learning Representations},
  volume={2024},
  pages={26516--26560},
  year={2024}
}

@inproceedings{li2025r,
  title={R*: Efficient reward design via reward structure evolution and parameter alignment optimization with large language models},
  author={Li, Pengyi and Jianye, HAO and Tang, Hongyao and Yuan, Yifu and Qiao, Jinbin and Dong, Zibin and Zheng, Yan},
  booktitle={Forty-second International Conference on Machine Learning},
  year={2025}
}

@article{sun2025large,
  title={A large language model-driven reward design framework via dynamic feedback for reinforcement learning},
  author={Sun, Shengjie and Liu, Runze and Lyu, Jiafei and Yang, Jing-Wen and Zhang, Liangpeng and Li, Xiu},
  journal={Knowledge-Based Systems},
  volume={326},
  pages={114065},
  year={2025},
  publisher={Elsevier}
}

@misc{fan2025forging,
  title={Forging Better Rewards: A Multi-Agent {LLM} Framework for Automated Reward Evolution},
  author={Haozhi Fan and Jiawei Du},
  year={2025},
  url={https://openreview.net/forum?id=Z6GStCfccl}
}

@article{gao2026rf,
  title={Rf-agent: automated reward function design via language agent tree search},
  author={Gao, Ning and Zhang, Xiuhui and Jiang, Xingyu and You, Mukang and Zhang, Mohan and Deng, Yue},
  journal={Advances in Neural Information Processing Systems},
  volume={38},
  pages={172532--172577},
  year={2026}
}

@article{wu2023read,
  title={Read and reap the rewards: Learning to play atari with the help of instruction manuals},
  author={Wu, Yue and Fan, Yewen and Liang, Paul Pu and Azaria, Amos and Li, Yuanzhi and Mitchell, Tom M},
  journal={Advances in Neural Information Processing Systems},
  volume={36},
  pages={1009--1023},
  year={2023}
}

@inproceedings{li2024auto,
  title={Auto mc-reward: Automated dense reward design with large language models for minecraft},
  author={Li, Hao and Yang, Xue and Wang, Zhaokai and Zhu, Xizhou and Zhou, Jie and Qiao, Yu and Wang, Xiaogang and Li, Hongsheng and Lu, Lewei and Dai, Jifeng},
  booktitle={Proceedings of the IEEE/CVF Conference on Computer Vision and Pattern Recognition},
  pages={16426--16435},
  year={2024}
}

@inproceedings{yifan2025llm,
  title={LLM Coach: Reward Shaping for Reinforcement Learning-Based Navigation Agent},
  author={Yifan, Hu and Bin-Bin, Hu and Bowen, Yuan and Hai-Tao, Zhang},
  booktitle={2025 Joint International Conference on Automation-Intelligence-Safety (ICAIS) \& International Symposium on Autonomous Systems (ISAS)},
  pages={1--6},
  year={2025},
  organization={IEEE}
}

@article{wei2026automated,
  title={An automated reinforcement learning reward design framework with large language model for cooperative platoon coordination},
  author={Wei, Dixiao and Yi, Peng and Lei, Jinlong and Hong, Yiguang and Dong, Hairong and Du, Yuchuan},
  journal={IEEE Transactions on Intelligent Transportation Systems},
  year={2026},
  publisher={IEEE}
}

@inproceedings{li2025llm,
  title={LLM-Assisted Semantically Diverse Teammate Generation for Efficient Multi-agent Coordination},
  author={Li, Lihe and Yuan, Lei and Liu, Pengsen and Jiang, Tao and Yu, Yang},
  booktitle={Forty-second International Conference on Machine Learning},
  year={2025}
}

@inproceedings{li2025remac,
  title={Re{MAC}: Large Language Model-Driven Reward Design for Multi-Agent Manipulation Collaboration},
  author={Pengyi Li and Hongyao Tang and Yifu Yuan and Jianye HAO},
  booktitle={Workshop on Scaling Environments for Agents},
  year={2025},
  url={https://openreview.net/forum?id=CWYWhLho0a}
}

@article{sun2025prof,
  title={PROF: An LLM-based Reward Code Preference Optimization Framework for Offline Imitation Learning},
  author={Sun, Shengjie and Lyu, Jiafei and Liu, Runze and Yan, Mengbei and Liu, Bo and Ye, Deheng and Li, Xiu},
  journal={arXiv preprint arXiv:2511.13765},
  year={2025}
}

@misc{makoviychuk2021isaacgymhighperformance,
  title={Isaac Gym: High Performance GPU-Based Physics Simulation For Robot Learning}, 
  author={Viktor Makoviychuk and Lukasz Wawrzyniak and Yunrong Guo and Michelle Lu and Kier Storey and Miles Macklin and David Hoeller and Nikita Rudin and Arthur Allshire and Ankur Handa and Gavriel State},
  year={2021},
  eprint={2108.10470},
  archivePrefix={arXiv},
  primaryClass={cs.RO},
  url={https://arxiv.org/abs/2108.10470}, 
}

@inproceedings{NEURIPS2022_217a2a38,
  author = {Chen, Yuanpei and Wu, Tianhao and Wang, Shengjie and Feng, Xidong and Jiang, Jiechuan and Lu, Zongqing and McAleer, Stephen and Dong, Hao and Zhu, Song-Chun and Yang, Yaodong},
  booktitle = {Advances in Neural Information Processing Systems},
  editor = {S. Koyejo and S. Mohamed and A. Agarwal and D. Belgrave and K. Cho and A. Oh},
  pages = {5150--5163},
  publisher = {Curran Associates, Inc.},
  title = {Towards Human-Level Bimanual Dexterous Manipulation with Reinforcement Learning},
  url = {https://proceedings.neurips.cc/paper_files/paper/2022/file/217a2a387f52c30755c37b0a73430291-Paper-Datasets_and_Benchmarks.pdf},
  volume = {35},
  year = {2022}
}

@article{xu2026deepseek,
  title={Deepseek-v4: Towards highly efficient million-token context intelligence},
  author={Xu, Anyi and Lin, Bangcai and Xue, Bing and Wang, Bingxuan and Xu, Bingzheng and Wu, Bochao and Zhang, Bowei and Lin, Chaofan and Dong, Chen and Ling, Chenchen and others},
  journal={arXiv preprint arXiv:2606.19348},
  year={2026}
}

@misc{openai2024gpt4o,
  title        = {Hello GPT-4o},
  author       = {{OpenAI}},
  year         = {2024},
  howpublished = {\url{https://openai.com/index/hello-gpt-4o/}},
}

@Book{Sutton+Barto:1998,
  author =       "Sutton, Richard S. and Barto, Andrew G.",
  title =        "Reinforcement Learning: An Introduction",
  publisher =    "MIT Press",
  year =         "1998",
  ISBN =         "0-262-19398-1",
  address =   "Cambridge, MA, USA",
  url = "http://www.cs.ualberta.ca/%7Esutton/book/ebook/the-book.html",
  bib2html_rescat = "Function Approximation, Partial Observability, Learning Methods, General RL, Applications",
}

@article{stanton2018deep,
  title={Deep curiosity search: Intra-life exploration improves performance on challenging deep reinforcement learning problems},
  author={Stanton, Christopher and Clune, Jeff},
  journal={arXiv preprint arXiv:1806.00553},
  year={2018}
}

@misc{hare2019dealingsparserewardsreinforcement,
  title={Dealing with Sparse Rewards in Reinforcement Learning}, 
  author={Joshua Hare},
  year={2019},
  eprint={1910.09281},
  archivePrefix={arXiv},
  primaryClass={cs.LG},
  url={https://arxiv.org/abs/1910.09281}, 
}

@Inbook{Eschmann2021,
  author="Eschmann, Jonas",
  title="Reward Function Design in Reinforcement Learning",
  bookTitle="Reinforcement Learning Algorithms: Analysis and Applications",
  year="2021",
  publisher="Springer International Publishing",
  address="Cham",
  pages="25--33",
  isbn="978-3-030-41188-6",
  doi="10.1007/978-3-030-41188-6_3",
  url="https://doi.org/10.1007/978-3-030-41188-6_3"
}
